\makeatletter\def\graphicscache@inhibit{true}\makeatother
\documentclass[conference]{IEEEtran}
\usepackage{times}

\usepackage[numbers]{natbib}
\usepackage{multicol}
\usepackage[bookmarks=true]{hyperref}

\usepackage[utf8]{inputenc}
\usepackage{graphicx}
\usepackage{tabularx}
\usepackage{threeparttable}
\usepackage{amsmath,amssymb}
\usepackage{bm}
\usepackage{xspace}
\usepackage[per-mode=symbol,binary-units=true,range-units=single,range-phrase=-,detect-weight=true,detect-family=true]{siunitx}
\usepackage{booktabs}
\usepackage{paralist}
\usepackage{subcaption}
\usepackage{url}
\usepackage{tikz}
\usepackage{pgfplots}
\usetikzlibrary{fit,shapes.callouts,shapes.geometric,backgrounds,positioning,arrows.meta,calc}
\graphicspath{{./figures/}}
\usepackage{adjustbox}
\usepackage{array}

\newcolumntype{R}[2]{%
    >{\adjustbox{angle=#1,lap=\width-(#2)}\bgroup}%
    l%
    <{\egroup}%
}
\newcommand*\rot{\multicolumn{1}{R{40}{1em}}}%

\newcommand{\ie}{i.e.,\ }
\newcommand{\eg}{e.g.,\ }
\newcommand{\cf}{cf.\xspace}

\newcommand{\reffig}[1]{Fig.~\ref{#1}}
\newcommand{\reftab}[1]{Tab.~\ref{#1}}
\newcommand{\refsec}[1]{Sec.~\ref{#1}}

\let\vec\bm
\newcommand{\norm}[1]{\left\lVert#1\right\rVert}

\DeclareSIUnit\pixel{P}

\begin{document}

\title{Real-Time Multi-View 3D Human Pose Estimation using Semantic Feedback to Smart Edge Sensors}

\author{\authorblockN{Simon Bultmann}
\authorblockA{Autonomous Intelligent Systems\\
University of Bonn, Germany\\
Email: {\tt bultmann@ais.uni-bonn.de}}
\and
\authorblockN{Sven Behnke}
\authorblockA{Autonomous Intelligent Systems\\
University of Bonn, Germany\\
Email: {\tt behnke@cs.uni-bonn.de}}
}

\maketitle

\begin{tikzpicture}[remember picture,overlay]
\node[anchor=north west,align=left,font=\sffamily,yshift=-0.2cm] at (current page.north west) {%
  Accepted for Robotics: Science and Systems (RSS), July 2021.
};
\end{tikzpicture}%

\begin{abstract}
We present a novel method for estimation of 3D human poses from a multi-camera setup, employing distributed smart edge sensors coupled with a backend through a semantic feedback loop.
2D joint detection for each camera view is performed locally on a dedicated embedded inference processor.
Only the semantic skeleton representation is transmitted over the network and raw images remain on the sensor board.
3D poses are recovered from 2D joints on a central backend, based on triangulation and a body model which incorporates prior knowledge of the human skeleton.
A feedback channel from backend to individual sensors is implemented on a semantic level.
The allocentric 3D pose is backprojected into the sensor views where it is fused with 2D joint detections.
The local semantic model on each sensor can thus be improved by incorporating global context information.
The whole pipeline is capable of real-time operation.
We evaluate our method on three public datasets, where we achieve state-of-the-art results and show the benefits of our feedback architecture, as well as in our own setup for multi-person experiments. Using the feedback signal improves the 2D joint detections and in turn the estimated 3D poses.
\end{abstract}

\IEEEpeerreviewmaketitle

\section{Introduction}
\label{sec:Introduction}
Accurate perception of humans is a challenging task with many applications in robotics and computer vision. It is a prerequisite \eg for safe navigation and anticipative movement of robots in the vicinity of people and can enable human-robot interaction or augmented reality scenarios.

\begin{figure}[t]
	\centering
	\begin{subfigure}[c]{0.49\linewidth}
		\centering
		\includegraphics[width=\linewidth]{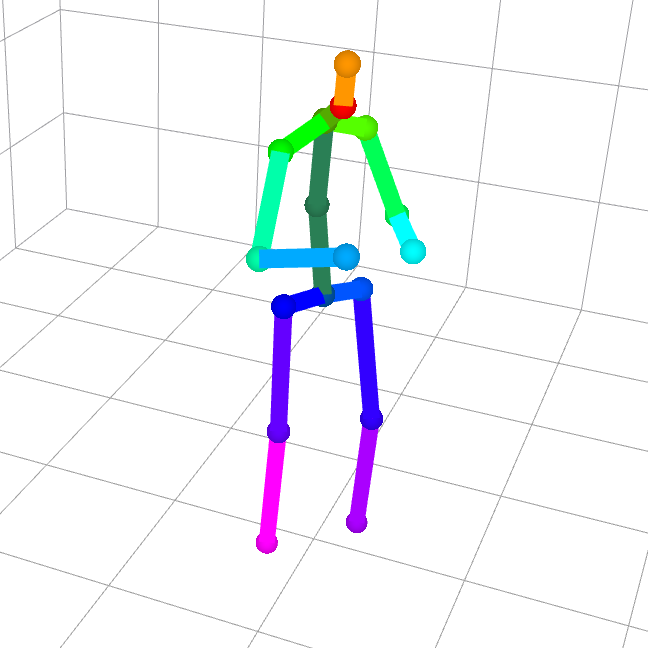}
		\label{fig:teaser1}
	\end{subfigure}
	\begin{subfigure}[c]{0.49\linewidth}
		\centering
		\includegraphics[width=\linewidth]{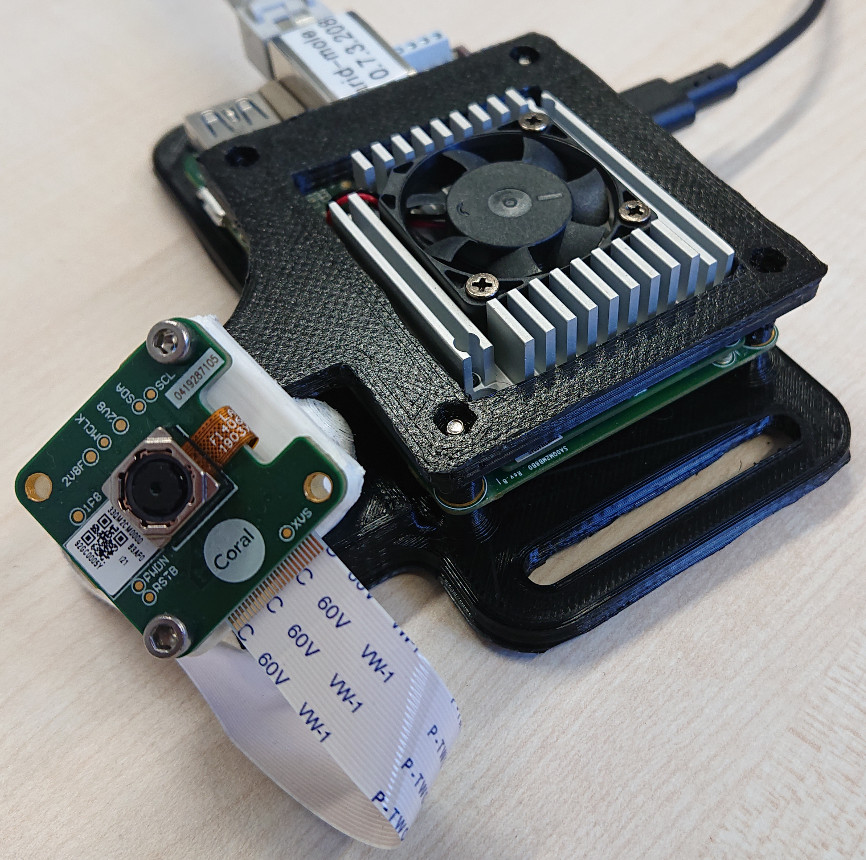}
		\label{fig:teaser2}
	\end{subfigure}
	\begin{subfigure}[c]{0.24\linewidth}
		\centering
		\includegraphics[width=\linewidth]{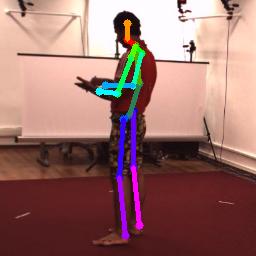}
		\label{fig:teaser31}
	\end{subfigure}
	\begin{subfigure}[c]{0.24\linewidth}
		\centering
		\includegraphics[width=\linewidth]{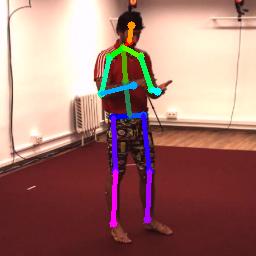}
		\label{fig:teaser32}
	\end{subfigure}
	\begin{subfigure}[c]{0.24\linewidth}
		\centering
		\includegraphics[width=\linewidth]{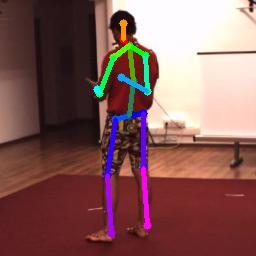}
		\label{fig:teaser33}
	\end{subfigure}
	\begin{subfigure}[c]{0.24\linewidth}
		\centering
		\includegraphics[width=\linewidth]{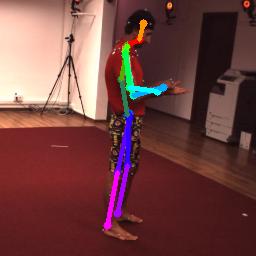}
		\label{fig:teaser34}
	\end{subfigure}
	
	\vspace{-0.7em}
	\caption{Multi-view 3D human pose estimation using smart edge sensors: Sensor board with attached camera (top-right). 2D pose detections from four views of the H3.6M data\-set~\cite{h36m} (bottom). Estimated 3D human skeleton (top-left).}
	\label{fig:teaser}
	\vspace{-1.2em}
\end{figure}

In this work, we address the task of 3D human pose estimation in allocentric world coordinates from a calibrated multi-camera setup.
Most state-of-the-art methods~\cite{qiu_cross_2019,pavlakos_harvesting_2017,chen_crossview_2020,remelli_lightweight_2020,dong_fast_2019} follow a two-step approach: First, 2D pose detections are generated for each available view (\cf\reffig{fig:teaser} bottom). Second, detections from multiple views are fused into a 3D human pose estimate and post-processed using a skeleton model (\cf\reffig{fig:teaser} top-left). Many recent methods focus more on accuracy than efficiency. They are thus difficult to employ in real-world scenarios with real-time constraints.

We propose a novel architecture for real-time multi-view 3D human pose estimation using distributed smart edge sensors for 2D pose estimation. Each camera view is interpreted locally using an embedded inference accelerator. The 2D human poses are streamed over a network to a central backend, where data association, triangulation and post-processing are performed to fuse the 2D detections into 3D skeletons. Furthermore, we propose a semantic feedback channel from backend to smart edge sensors. The allocentric 3D pose estimate is backprojected into the respective local views where it is combined with the joint detections. Thus, global context information can be incorporated into the local 2D semantic model of the individual smart edge sensor, which improves the pose estimation result.

The use of distributed smart edge sensors has several advantages over the centralized approaches more common in literature. As the images are processed directly on the sensor boards, raw images are not sent to the backend and only the 2D pose information has to be transmitted over the network. This significantly reduces the required communication bandwidth and furthermore mitigates privacy issues, as the abstract semantic information contains no personal details. Moreover, using a dedicated inference accelerator for each camera lessens hardware requirements on the backend side, which, in a centralized architecture, can quickly become the bottleneck.
On the other hand, using an embedded sensor platform poses challenges, as the employed vision models need to meet the limitations of the hardware. For this, we propose a lightweight 2D pose estimation model for efficient image processing locally on the sensors, on the edge of the network.

In summary, the main contributions of our work are:
\begin{itemize}
\item a new real-time method for multi-view 3D human pose estimation dividing the computation between smart edge sensors performing image analysis locally for each camera view and a backend fusing the semantic interpretations of individual views and using a computationally efficient skeleton model to incorporate prior knowledge,
\item a novel 3D\,/\,2D feedback architecture enabling bidirectional communication on a semantic level between sensors and backend, and
\item an extensive evaluation of the proposed approach on the single-person H3.6M dataset~\cite{h36m}, the multi-person Campus and Shelf datasets~\cite{belagiannis_3d_2014}, and in own multi-person experiments in a less-controlled environment.
\end{itemize}
 
\section{Related Work}
\label{sec:Related_Work}
Human pose estimation from multi-camera input has been investigated for many years in the computer vision and robotics communities.
Early works~\cite{belagiannis_3d_2014,belagiannis_multiple_2015,burenius_3d_2013} use manually designed image features, such as HOG descriptors~\cite{dalal2005histograms}, for 2D part detection and combine multiple views using a graph-based body model.
With the increasing success of deep-learning methods, more recent approaches~\cite{qiu_cross_2019,pavlakos_harvesting_2017,dong_fast_2019} employ 2D convolutional neural networks (CNNs) for human joint detection~\cite{cao_openpose_2018,xiao_simple_2018} and recover the 3D pose using variants of the Pictorial Structures Model (PSM)~\cite{belagiannis_3d_2014,burenius_3d_2013}. In these approaches, the body model consists of a graph with 3D joint locations as nodes and pairwise articulation constraints on the edges. While the PSM body model recovers 3D poses accurately, it is computationally very expensive and generally not real-time capable, due to a large volumetric grid used as discrete state space for optimization.
In our work, we also employ a graph-based body model but use a fast iterative optimization scheme~\cite{kaess2012isam2}, achieving real-time operation.

Qiu et al.~\cite{qiu_cross_2019} present an approach for cross-view fusion to improve the estimated 2D poses of individual camera views. 3D poses are recovered using an offline recursive PSM implementation with a processing time of several seconds per frame~\cite{remelli_lightweight_2020}.
In our work, we take up the idea of across-sensor viewpoint fusion but propose a different formulation. Qiu et al.~\cite{qiu_cross_2019} implement the fusion between perspectives on a purely 2D basis, using epipolar constraints. Hence, a 2D joint in one view will be associated with all features on the corresponding epipolar lines of other views, which can be ambiguous.
In contrast, we implement a semantic 3D\,/\,2D feedback channel from backend to sensors based on reprojection of the estimated 3D skeleton into the individual camera views.

Several recent methods with a focus on computational efficiency have been proposed~\cite{chen_crossview_2020,remelli_lightweight_2020}. In these approaches, 3D pose estimation is based on direct triangulation of 2D joint detections without usage of an expensive body model. Chen et al.~\cite{chen_crossview_2020} propose a fast iterative triangulation scheme but assume 2D pose detections as given. Remelli et al.~\cite{remelli_lightweight_2020} consider the whole pipeline including 2D keypoint estimation but use a fully centralized approach while our method employs distributed sensors for 2D pose estimation.

Naikal et al.~\cite{naikal_joint_2014} proposed a system for human joint detection and action recognition using a network of smart cameras transmitting only abstract image features to a central processing station. However, at the time, no CNN-based vision models were available for pose estimation on mobile devices, limiting the performance of their framework. Furthermore, their communication channel is purely feed-forward---no feedback for viewpoint fusion is implemented.

Research interest in computer vision models that run efficiently also on mobile and embedded devices has significantly increased in recent years.
MobiPose~\cite{zhang2020mobipose} investigates human pose estimation on smartphone SoCs without dedicated inference accelerators, using motion vector-based tracking. Xiao et al.~\cite{xiao_simple_2018} propose a simple CNN architecture consisting only of a feature extractor and a deconvolutional head but use a standard ResNet backbone~\cite{he_deep_2016}.
Popular lightweight backbone architectures include EfficientNet~\cite{tan_efficientnet_2019} and MobileNets~\cite{mobilenet_2017,mobilenetv32019}.
These architectures greatly reduce the number of parameters w.r.t. standard CNN feature extractors like ResNet, \eg by replacing convolutions with depthwise-separable convolutions.
Moreover, tensor processors for inference acceleration, like the Google Edge\,TPU~\cite{yazdanbakhsh_edgetpu_2021}, can be employed to efficiently run a CNN vision model within a limited size and energy budget. For compatibility with the Edge\,TPU, weights and activations of the model need to be quantized to 8-bit integer values using a quantization scheme as proposed by Jacob et al.~\cite{quantization_2018}.

To the best of our knowledge, the proposed framework is the first approach for real-time 3D human pose estimation using multiple smart edge sensors which perform 2D pose estimation on-device and incorporate global context via semantic feedback from the backend. 
\section{Method}
\label{sec:Method}
\begin{figure*}[!ht]
  \centering
  \vspace*{0.6em}
  \resizebox{0.9\linewidth}{!}{%
\begin{tikzpicture}[font=\sffamily,on grid,>={Stealth[inset=0pt,length=4pt,angle'=45]}]
\tikzset{every node/.append style={node distance=3.0cm}}
\tikzset{img_node/.append style={minimum size=1.5em,minimum height=3em,align=center,scale=0.65}}
\tikzset{content_node/.append style={minimum size=1.5em,minimum height=3em,minimum width={width("Triangulation") +1.0em},draw,align=center,scale=0.65,fill=blue!15!white}}
\tikzset{label_node/.append style={scale=0.5, near start}}
\tikzset{junction/.append style={circle, fill=black, minimum size=3pt, draw}}

\definecolor{red}{rgb}     {0.9,0.0,0.0}
\definecolor{green}{rgb}   {0.0,0.5,0.0}
\definecolor{blue}{rgb}    {0.0,0.0,0.5}
\definecolor{grey}{rgb}    {0.5,0.5,0.5}

\draw[thick, rounded corners, grey!20!white,fill] (-0.9, -0.55) -- (2.9, -0.55) -- (2.9, 0.55) -- (-0.9, 0.55) -- cycle;
\draw[thick, rounded corners, grey!20!white,fill] (-0.9, -0.55 - 1.95) -- (2.9, -0.55 - 1.95) -- (2.9, 0.55 - 1.95) -- (-0.9, 0.55 - 1.95) -- cycle;
\draw[thick, rounded corners, grey!20!white,fill] (4.4, 0.6) -- (8.7, 0.6) -- (8.7, -2.95) -- (4.4, -2.95) -- cycle;

\node(camera1)[img_node] at (0, 0) {\includegraphics[width=1.6cm]{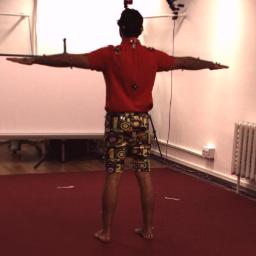}};
\node(posedet1)[content_node, right of=camera1]{2D Pose\\Estimation};
\node(skel1)[img_node, above right=-0.1cm and 1.8cm of posedet1] {2D Pose\\[4pt]
        \setlength{\fboxsep}{0.5pt}%
        \setlength{\fboxrule}{0.5pt}%
        \fbox{\includegraphics[width=1.3cm]{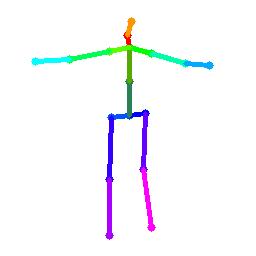}}};

\node(cameraN)[img_node, below of=camera1] {\includegraphics[width=1.6cm]{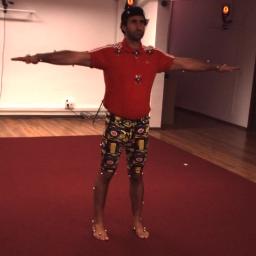}};
\node(posedetN)[content_node, right of=cameraN]{2D Pose\\Estimation};
\node(skelN)[img_node, above right=0.24cm and 1.8cm of posedetN] {\setlength{\fboxsep}{0.5pt}%
        \setlength{\fboxrule}{0.5pt}%
        \fbox{\includegraphics[width=1.3cm]{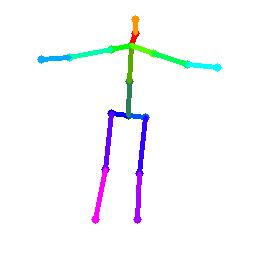}}};
        
\node(posedet_dummy) at ($(posedet1)!0.5!(posedetN)$){};

\node(triang)[content_node, above right=0.0cm and 3.5cm of posedet_dummy]{Multi-View\\Triangulation};
\node(model)[content_node, right of=triang]{3D Skeleton\\Model};
\node(prior)[content_node, above right=1cm and 0cm of model]{Bone-Length\\Prior};
\node(pred)[content_node, above right=-1.5cm and 0cm of model]{Prediction};
\node(reproj)[content_node, left of=pred]{Reprojection};
\node(junct)[junction, above right=0.0cm and 1.1cm of model, scale=0.3]{};
\node(out)[above right=0.0cm and 2.5cm of model, scale=0.65, align=center]{\includegraphics[width=2.6cm]{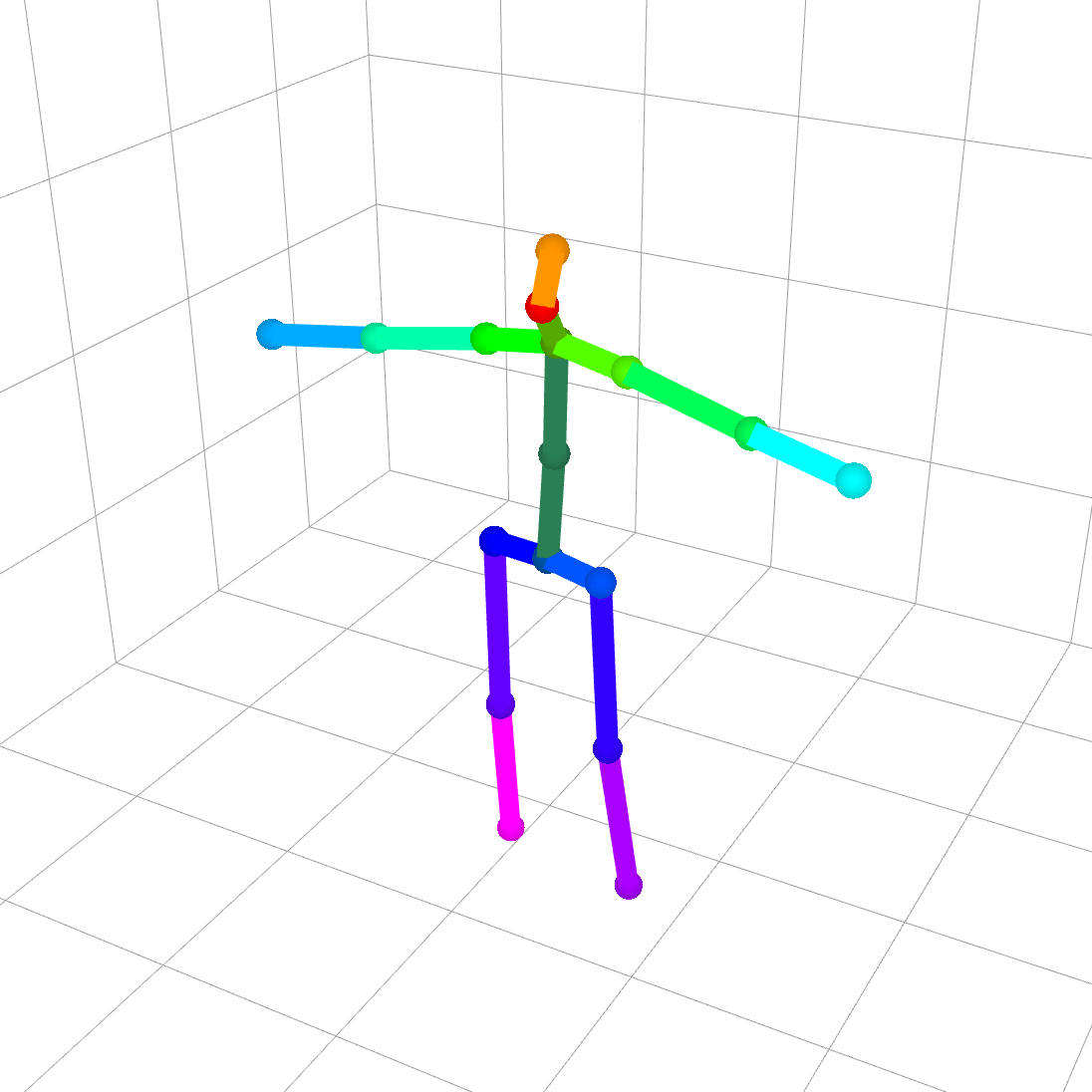}\\3D Pose};

\draw[->, thick] (camera1) -- node[label_node,midway,above] {RGB} node[label_node,midway,below] {Image} (posedet1);
\draw[->, thick] (cameraN) -- node[label_node,midway,above] {RGB} node[label_node,midway,below] {Image} (posedetN);

\draw[->, thick] (posedet1) -| ++(1.2, -0.5) |- ([yshift= 2mm]triang.west);
\draw[->, thick] (posedetN) -| ++(1.2,  0.5) |- ([yshift=-2mm]triang.west);
\draw[->, dashed, thick] (posedet_dummy) + (1.0,0) -- (triang);
\draw[dotted, thick] ([yshift=-6.7mm]$(camera1)!0.5!(posedet1)$) -| ([yshift=8mm]$(cameraN)!0.5!(posedetN)$);

\draw[->, thick] (triang) -- (model);
\draw[thick] (model) -- (junct);
\draw[->, thick] (junct) -- (out);

\draw[->, thick] (prior) -- (model);
\draw[->, thick] (junct) |- (pred);
\draw[->, thick] (pred) -- (reproj);

\draw[->, thick, red] ([yshift= 2mm]reproj.west) -| ++(-1.7, 1.0) -| (posedet1);
\draw[->, thick, red] ([yshift=-2mm]reproj.west) -| node[label_node,below] {\textbf{Semantic Feedback}} (posedetN);
\draw[->, dashed, thick, red] (reproj.west) -- ++(-1.5, 0.0);

\node[scale=0.6, anchor=north west] at (-0.9, 0.85) {\textbf{Smart Edge Sensor $1$}};
\node[scale=0.6, anchor=north west] at (-0.9, 0.85 - 1.95) {\textbf{Smart Edge Sensor $N$}};
\node[scale=0.6, anchor=north west] at (4.45, 0.55) {\textbf{Backend}};

\end{tikzpicture}

}
  \caption{Overview of the proposed pipeline for 3D human pose estimation using smart edge sensors and semantic feedback. Images are analyzed locally on the sensor boards. Semantic pose information is transmitted to the backend where multiple views are fused into a 3D skeleton. The 3D pose is reprojected into local views and sent to sensors as semantic feedback. }
  \label{fig:pipeline}
  \vspace{-1.2em}
\end{figure*}
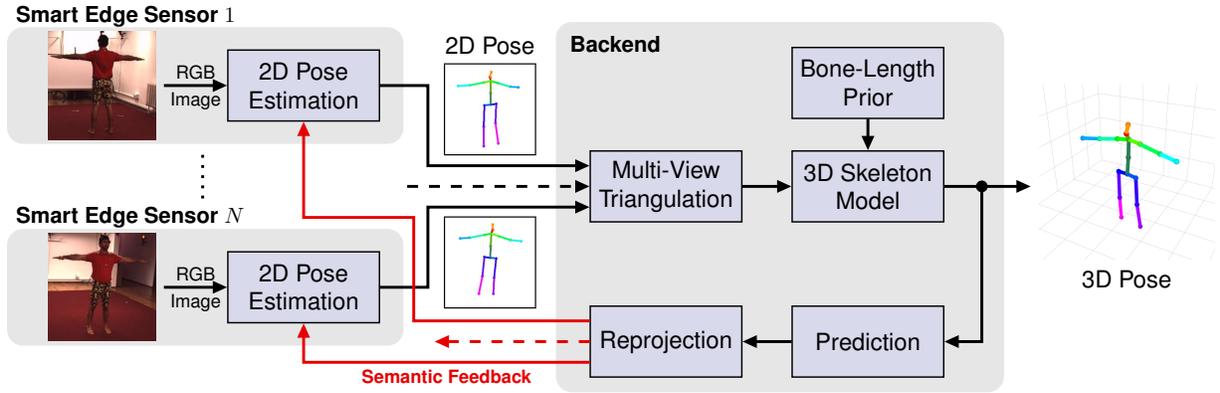
An overview of our proposed approach is given in \reffig{fig:pipeline}. We consider scenarios where $N$ calibrated cameras with known projection matrices $\vec{P}_i$ perceive a scene with one or several individuals from multiple viewpoints. Our method is described for the single person case in the following. Extensions to handle multiple persons are described in \refsec{sec:multiperson}.

2D locations of a fixed set of $J$ human joints $\{\vec{u}_i^j\}_{j=1}^J$ in camera view $i$, corresponding confidence values $c_i^j$ and covariance matrices $\vec{\Sigma}_i^j$ are calculated directly on the respective smart edge sensor board using the vision model described in \refsec{sec:pose2d}. The 2D pose information is then transmitted over a network to a central backend, using the robot operation system (ROS)~\cite{ros_quigley_09} as middleware for communication.

The clocks of sensors and backend are soft\-ware-syn\-chro\-nized and each 2D pose message includes a timestamp representing the capture time of the corresponding image.
Sets of $N$ corresponding messages, one for each view, are determined based on these timestamps, and raw 3D poses are recovered via triangulation as detailed in \refsec{sec:triang}.

A skeleton model (cf. \refsec{sec:skeleton_model}), incorporating prior information on the typical bone-lengths of the human skeleton, is then applied and outputs the final estimated 3D pose.

A semantic feedback channel from backend to sensors is implemented as described in \refsec{sec:feedback}, which enables each individual view to benefit from the fused 3D information. For this, first, a prediction step is performed to compensate for the pipeline delay. Second, the predicted 3D skeleton is reprojected into each camera view and sent to the sensors where it is incorporated into the local 2D pose estimation.

\subsection{2D Human Pose Estimation on Smart Edge Sensor}
\label{sec:pose2d}
\begin{figure}[t]
	\vspace*{0.7em}
	\centering
	\begin{subfigure}[c]{0.22\linewidth}
		\centering
		\includegraphics[trim=50px 50px 50px 50px,clip,width=\linewidth]{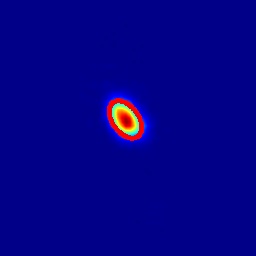}
	\end{subfigure}
	\begin{subfigure}[c]{0.22\linewidth}
		\centering
		\includegraphics[trim=40px 75px 60px 25px,clip,width=\linewidth]{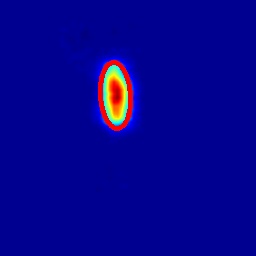}
	\end{subfigure}
	\begin{subfigure}[c]{0.22\linewidth}
		\centering
		\includegraphics[trim=50px 60px 50px 40px,clip,width=\linewidth]{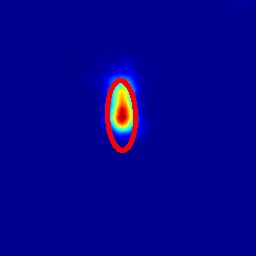}
	\end{subfigure}
	\begin{subfigure}[c]{0.22\linewidth}
		\centering
		\includegraphics[trim=30px 80px 70px 20px,clip,width=\linewidth]{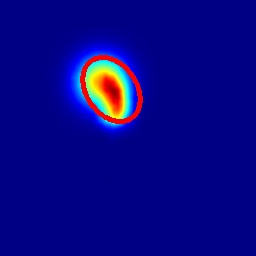}
	\end{subfigure}
	\caption{Heatmaps and derived covariances ($3\sigma$ ellipses).}
	\label{fig:hm_cov}
	\vspace{-1.5em}
\end{figure}
The smart edge sensor platform employed in this work (cf. \reffig{fig:teaser} top-right) is based on the Google Edge\,TPU Dev Board~\cite{edgetpu_devboard}, equipped with an ARM Cortex-A53 quad-core processor, the Edge\,TPU inference accelerator and \SI{1}{\giga\byte} of shared RAM. A \SI{5}{\mega\pixel} RGB camera is connected to the board via the MIPI-CSI2 interface.

We adopt the CNN architecture of Xiao et al.~\cite{xiao_simple_2018} for 2D human pose estimation, consisting of a backbone feature extractor and three transposed convolution layers to extract heatmaps from image features. To achieve real-time performance on the mobile sensor platform, we exchange the ResNet backbone used by Xiao et al.~\cite{xiao_simple_2018} with the significantly more lightweight MobileNetV3 feature extractor~\cite{mobilenetv32019}. Furthermore, for execution on the Edge\,TPU, the model is quantized for 8-bit integer inference using post-training quantization~\cite{quantization_2018} as implemented in the TensorFlow ML framework~\cite{abadi2016tensorflow}.
In multi-person scenarios, a detector is also run on the sensor boards to provide person crops for the pose estimation network. It is based on the Single Shot Detector (SSD) architecture~\cite{liu_ssd_2016}, also using the MobileNetV3 backbone.

The output heatmaps $\vec{H}_\text{det}$ of the pose estimation model are a multi-channel image with one channel per joint, encoding the confidence of a joint being present at the pixel location. 2D joint locations $\vec{u}^j = [u^j, v^j]^\text{T}$ are inferred as global maxima of the resp. heatmap channel, as single person crops are processed. The value of the heatmap at the joint position gives the corresponding confidence $c^j$. Only joints with confidence above a minimum threshold are considered as valid detections.

The covariance matrices $\vec{\Sigma}^j$ are determined as proposed by Pasqualetto et at.~\cite{pasqualetto2020cnn}: Heatmap pixels with values above a threshold contribute to the empirical covariance with their $x$- and $y$-locations, weighted by the respective confidence:
\begin{align}
\vec{\Sigma}^j &= \begin{bmatrix} \sigma^2_{xx} & \sigma^2_{xy} \\ \sigma^2_{xy} & \sigma^2_{yy} \end{bmatrix} \,,\\
\sigma^2_{xy} &= \frac{1}{K} \sum_{k=1}^K c^j_k \left(x_k - u^j\right) \cdot \left(y_k - v^j \right)\,,
\end{align}
where $K$ is the number of contributing pixels and the mean is replaced by the peak $\vec{u}^j$ to model a distribution about the detected 2D joint location.
Some representative examples of heatmaps and extracted covariances are shown in \reffig{fig:hm_cov}. The uncertainty in the heatmaps, including their directionality, is well captured by the covariance ellipses. Note, that the dispersion for asymmetric heatmaps, as in the third example of \reffig{fig:hm_cov}, is overestimated by the proposed procedure.

\subsection{Multi-View Fusion}
\label{sec:triang}
The 3D position $\hat{\vec{x}}^j$ of each joint $j$ is recovered from a set of 2D detections $\{\vec{u}_i^j\}_{i=1}^N$ via triangulation using the Direct Linear Transform (DLT)~\cite{hartley2003multiple}. The relationship between 2D points $\vec{u}_i = [u_i, v_i]^\text{T}$ from camera view $i\in\{1,\ldots, N\}$ and 3D point $\hat{\vec{x}}$ can be written as:
\begin{align}
\vec{A}\tilde{\vec{x}} = \vec{0}\,,\label{eqn:dlt}
\end{align}
with
\begin{align}
\vec{A} = \begin{bmatrix}
	u_1\vec{p}_{1, 3}^{\text{T}} - \vec{p}_{1, 1}^{\text{T}}\\
	v_1\vec{p}_{1, 3}^{\text{T}} - \vec{p}_{1, 2}^{\text{T}}\\
	\cdots \\
	u_N\vec{p}_{N, 3}^{\text{T}} - \vec{p}_{N, 1}^{\text{T}}\\
	v_N\vec{p}_{N, 3}^{\text{T}} - \vec{p}_{N, 2}^{\text{T}}
\end{bmatrix} \in \mathbb{R}^{2N\times4}\,,
\end{align}
where $\tilde{\vec{x}} \in \mathbb{R}^4$ are the homogeneous coordinates of $\hat{\vec{x}}$ and $\vec{p}_{i,k}^{\text{T}}$ denotes the $k$-th row of projection matrix $\vec{P}_i \in \mathbb{R}^{3\times4}$. According to the DLT algorithm~\cite{hartley2003multiple}, \eqref{eqn:dlt} is solved by a singular value decomposition (SVD) on $\vec{A}$, taking the unit singular vector corresponding to the smallest singular value of $\vec{A}$ as solution for $\tilde{\vec{x}}$. Finally, $\tilde{\vec{x}}$ is divided by its fourth coordinate to obtain the 3D vector $\hat{\vec{x}} = \tilde{\vec{x}}/(\tilde{\vec{x}})_4$.

The above formulation \eqref{eqn:dlt} assumes that all 2D detections make a similar contribution to the triangulation. However, 2D joint positions cannot be estimated reliably on some views, \eg due to occlusions, which in turn degrades the result.
The reliability of a detection is expressed by the heatmap confidence value $c_i$ and can be incorporated into the DLT by multiplying each row of $\vec{A}$ with the corresponding element of a weight vector $\vec{w}$, reformulating \eqref{eqn:dlt} as:
\begin{align}
\left(\vec{w}\circ\vec{A}\right)\tilde{\vec{x}} = \vec{0}\,,\label{eqn:dlt_weighted}
\end{align}
with
\begin{align}
\vec{w}=\left(\frac{c_1}{\norm{\vec{a}_{1}^\text{T}}}, \frac{c_1}{\norm{\vec{a}_2^\text{T}}}, \ldots, \frac{c_N}{\norm{\vec{a}_{2N-1}^\text{T}}}, \frac{c_N}{\norm{\vec{a}_{2N}^\text{T}}}\right)
\end{align}
and $\circ$ the Hadamard product, similar to the approach of Chen et al.~\cite{chen_crossview_2020}. The confidence values in $\vec{w}$ are divided by the $\text{L}^2$-norm of the corresponding row of $\vec{A}$ to compensate for the different image locations of the joints in each view.

To obtain the 3D joint position $\hat{\vec{x}}^j$ and its covariance $\hat{\vec{\Sigma}}^j_\text{3D}$, deterministic samples are propagated through the triangulation according to the Unscented Transform~\cite{julier2004unscented}. Sigma points are generated from the mean vector $\vec{\mu}^j= [\vec{u}_1^{j\text{T}},\ldots,\vec{u}_N^{j\text{T}}]^\text{T}$ and the block-diagonal matrix containing the 2D covariances $\vec{\Sigma}^j_i$ extracted from each heatmap. Each set of samples is triangulated according to \eqref{eqn:dlt_weighted} and $\hat{\vec{x}}^j$ and $\hat{\vec{\Sigma}}^j_\text{3D}$ are determined as sample mean and covariance of the resulting points using weights given by the Unscented Transform.

\subsection{Skeleton Model}
\label{sec:skeleton_model}
We employ a factor graph model~\cite{kaess2012isam2} representing the tree structure of the human body, with 3D joint positions $\vec{x}^j$ as nodes connected by unary and pairwise factors on the edges.

The unary constraints are given by the triangulated joint positions $\hat{\vec{x}}^j$ and covariances $\hat{\vec{\Sigma}}^j_\text{3D}$ and follow a 3D Gaussian noise model:
\begin{align}
f\left(\vec{x}^j\right) \sim \mathcal{N}\left(\vec{x}^j |\, \hat{\vec{x}}^j, \hat{\vec{\Sigma}}^j_\text{3D}\right)\,.
\end{align}

The pairwise factors model typical limb-lengths of the human body and also follow a Gaussian noise model:
\begin{align}
g\left(\vec{x}^j, \vec{x}^k\right) \sim \mathcal{N}\left(\norm{\vec{x}^j - \vec{x}^k} |\, l_{j,k}, \sigma_\text{l}\right)\,,
\end{align}
with $\norm{\vec{x}^j - \vec{x}^k}$ the Euclidean distance between joints $\vec{x}^j$ and $\vec{x}^k$. $l_{j,k}$ and $\sigma_\text{l}$ denote mean and standard deviation of the length of the corresponding limb determined from the statistics of the H3.6M dataset~\cite{h36m}.

The final 3D human poses are obtained by optimizing the factor graph using the Levenberg-Marquardt algorithm and the \textit{gtsam}-framework~\cite{kaess2012isam2}. The optimization is initialized with the poses from the previous frame, predicted using a linear velocity model.

\subsection{Semantic Feedback}
\label{sec:feedback}
To enable the local semantic models of each sensor to benefit from the globally fused 3D pose, a feedback channel from backend to sensors is implemented in our framework.

First, the motion of the 3D skeleton is predicted using a linear velocity model for each joint to compensate for the pipeline delay $\Delta t$.
Second, predicted 2D joint positions $\{\hat{\vec{u}}_i^j\}$ and their image-plane covariances $\{\hat{\vec{\Sigma}}_i^j\}$ are determined by reprojecting the predicted 3D pose and its covariance extracted from the factor graph into each sensor view $i$ using the projection matrix $\vec{P}_i$ and the Unscented Transform~\cite{julier2004unscented}.

The reprojected feedback skeleton is sent to the smart edge sensors, where a feedback heatmap $\vec{H}_\text{fb}$ is rendered to be fused with the detected heatmap $\vec{H}_\text{det}$ of the current image crop. For each joint $\hat{\vec{u}}^j$, a 2D Gaussian blob is rendered in the corresponding heatmap channel according to the reprojected covariance matrix $\hat{\vec{\Sigma}}^j$.

The heatmaps are fused via weighted addition of detection, feedback, and their element-wise multiplication:
\begin{align}
\vec{H}_\text{fused} = s \left(\left(1\!-\!\alpha\!-\!\beta\right)\vec{H}_\text{det} + \alpha \vec{H}_\text{fb} + \beta\left(\vec{H}_\text{fb}\circ\vec{H}_\text{det}\right)\right),\label{eqn:hm_fusion}
\end{align}
with $\alpha + \beta < 1$. The scale $s$ is set as $(1-\alpha-\beta)^{-1}$ to ensure that positive feedback always increases the joint confidence. The feedback gains $\alpha$ and $\beta$ are important design parameters of our method. A sufficient weight must be accorded to the feedback to improve the raw detections, but too high gains can cause instability.
Hence, the feedback gains are learned using a hyper-parameter search~\cite{bergstra2013hyperopt} optimizing the 3D pose error.

The above formulation models an arbitrary combination of additive and multiplicative feedback and can efficiently be executed on the embedded processor of the sensor board. 
Examples of the heatmap fusion are shown in \reffig{fig:feedback}.
Through the feedback loop, evidence for joint occurrence from detection and feedback is combined in the fused heatmap, improving the accuracy of the joint locations and reducing their uncertainty.
\begin{table}[t]
\vspace*{0.7em}
\caption{2D Joint Detection Rate (JDR) ($\%$) for different joint classes, feedback modes and training data.}
\label{tab:error_2d_joint}
\centering
\setlength{\tabcolsep}{2.5pt}
\begin{threeparttable}
\begin{tabular}{lc|ccccccc}
  \toprule
   Feedback & Training Data & Hips & Knees & Ankls & Shlds & Elbs & Wrists & \textbf{Avg}\\
  \midrule
   w/o fb   & H3.6M      	& 99.2  & 96.1  & 90.3  & 93.3  & 93.3 & 89.1 & 95.1  \\
   w fb    & H3.6M      	& \textbf{99.5}  & 97.6  & 96.1  & 97.2  & 96.5 & \textbf{94.8} & 97.5  \\ %
   w/o fb   & COCO + H3.6M & 99.3  & 97.1  & 96.9  & 98.9  & 96.2 & 92.8 & 97.6  \\
   w fb     & COCO + H3.6M & 99.3  & \textbf{98.0}  & \textbf{97.8}  & \textbf{99.0}  & \textbf{97.1} & \textbf{94.8} & \textbf{98.2}  \\
  \bottomrule
\end{tabular}
\end{threeparttable}
\vspace{-.5em}
\end{table}
\begin{table}[t]
\caption{3D pose error (\textit{mm}) for different joint classes, feedback modes and training data.}
\label{tab:error_3d_joint}
\centering
\setlength{\tabcolsep}{2.5pt}
\begin{threeparttable}
\begin{tabular}{lc|ccccccc}
  \toprule
   Feedback & Training Data & Hips & Knees & Ankls & Shlds & Elbs & Wrists & \textbf{Avg} \\
  \midrule
   w/o fb   & H3.6M         & 22.2 & 29.4  & 58.6  & 40.5  & 43.8 & 39.8   & 32.9 \\
   w fb     & H3.6M         & 22.1 & 28.0  & 47.2  & 36.7  & 38.6 & 33.9   & 29.8 \\
   w/o fb   & COCO + H3.6M  & \textbf{19.2} & 25.5  & 38.0  & 25.6  & 30.7 & 29.4   & 24.0 \\
   w fb     & COCO + H3.6M  & \textbf{19.2} & \textbf{24.9}  & \textbf{36.9}  & \textbf{25.5}  & \textbf{29.9} & \textbf{28.3}   & \textbf{23.5} \\
  \bottomrule
\end{tabular}
\end{threeparttable}
\vspace{-1.5em}
\end{table}
\begin{table*}[!ht]
\vspace*{0.7em}
\caption{Evaluation result on H3.6M dataset: MPJPE 3D pose error (\textit{mm}) per action type. 3D poses after application of the resp. post-processing or skeleton model are reported. $^+$ denotes using additional training data.}
\label{tab:error_3d_act}
\centering
\setlength{\tabcolsep}{4.9pt}
\begin{threeparttable}
\begin{tabular}{l|cccccccccccccccc}
  \toprule
   Method 												& Dir. & Disc. & Eat & Greet & Phone & Photo & Pose & Purch. & Sit & SitD. & Smoke & Wait & WalkD. & Walk & WalkT. & \textbf{Avg}\\
   \midrule
   \citet{pavlakos_harvesting_2017}    & 41.2 & 49.2  & 42.8 & 43.4  & 55.6  & 46.9  & 40.3 & 63.7   & 97.6 &119.9 & 52.1  & 42.7 & 51.9   & 41.8 & 39.4   & 56.9 \\
   \citet{tome_rethinking_2018}			   & 43.3 & 49.6  & 42.0 & 48.8  & 51.1  & 64.3  & 40.3 & 43.3   & 66.0 & 95.2 & 50.2  & 52.2 & 51.1   & 43.9 & 45.3   & 52.8 \\
   \citet{qiu_cross_2019}					   & 28.9 & 32.5  & 26.6 & 28.1  & \textbf{28.3}  & 29.3  & 28.0 & 36.8   & 42.0 & \textbf{30.5} & 35.6  & 30.0 & 28.3   & 30.0 & 30.5   & 31.2 \\
   \citet{remelli_lightweight_2020}     & 27.3 & 32.1  & \textbf{25.0} & \textbf{26.5}  & 29.3  & 35.4  & 28.8 & 31.6   & \textbf{36.4} & 31.7 & 31.2  & 29.9 & \textbf{26.9}   & 33.7 & 30.4   & 30.2 \\
   \midrule
   Ours, w/o fb										   & 27.7 & 36.5  & 27.8 & 27.1  & 33.9  & 33.1 & 29.3  & 33.6   & 41.3 & 42.5 & 32.8  & 33.5 & 33.3   & 27.8   & 27.2 & 32.9 \\
   Ours, w fb										   & \textbf{27.1} & \textbf{29.9}  & 27.0 & \textbf{26.5}  & 31.3  & \textbf{28.9} & \textbf{27.1}  & \textbf{29.8}   & 36.5 & 36.0 & \textbf{30.8}  & \textbf{29.3} & 29.7   & \textbf{27.3}   & \textbf{26.3} & \textbf{29.8} \\
   \midrule\midrule
   \citet{qiu_cross_2019}$^+$				   & 24.0 & 26.7  & 23.2 & 24.3  & 24.8  & \textbf{22.8}  & 24.1 & 28.6   & 32.1 & 26.9 & 31.0  & 25.6 & 25.0   & 28.1 & 24.4   & 26.2 \\
   Ours$^+$, w/o fb								       & \textbf{22.4} & 24.3  & 22.4 & \textbf{21.7}  & 24.6  & 24.7 & 22.4  & \textbf{22.6}   & 26.8 & 28.4 & 25.0  & 23.1 & \textbf{24.5}   & 22.0   & 21.5 & 24.0 \\
   Ours$^+$, w fb									   & \textbf{22.4} & \textbf{24.0}  & \textbf{22.2} & \textbf{21.7}  & \textbf{24.0}  & 23.9 & \textbf{22.1}  & \textbf{22.6}   & \textbf{26.0} & \textbf{26.8} & \textbf{24.5}  & \textbf{22.8} & 24.6  & \textbf{21.8}   & \textbf{21.3} & \textbf{23.5} \\
  \bottomrule
\end{tabular}
\end{threeparttable}
\vspace{-1em}
\end{table*}

\subsection{Multi-Person Pose Estimation}
\label{sec:multiperson}
To handle real-world scenes (cf. \refsec{sec:eval_campus} and \ref{sec:eval_realworld}), we extend our method to estimate the poses of multiple persons at a time.
Person detections are associated across camera views based on the epipolar distance of their joints using the efficient iterative greedy matching proposed by Tanke et al.~\cite{tanke_iterative_2019}. The rest of the pipeline is then run for each person observed in at least two views to compute 3D poses and feedback.
A feedback skeleton is associated to its corresponding 2D detection based on the IoU overlap of their bounding boxes. 
\section{Evaluation and Experiments}
\label{sec:Evaluation}
We evaluate the proposed approach on three widely-used public datasets: The Human 3.6M dataset~\cite{h36m} and the multi-person Campus and Shelf datasets~\cite{belagiannis_3d_2014}, as well as on own data from experiments in our lab. We make our C++ implementation publicly available at \href{https://github.com/AIS-Bonn/SmartEdgeSensor3DHumanPose}{https://github.com/AIS-Bonn/SmartEdgeSensor3DHumanPose}.
\subsection{Dataset and Metrics}
\subsubsection{Human 3.6M}
The Human 3.6M dataset~\cite{h36m} is a large-scale public dataset for single-person multi-view 3D human pose estimation. It contains 3.6 million frames of 11 different actors, captured by four synchronized cameras together with ground truth 2D and 3D poses.

We measure the 2D pose estimation accuracy as the percentage of correctly detected joints, the Joint Detection Rate (JDR). A joint is correctly detected when its distance towards the corresponding ground-truth annotation is smaller than a threshold. We set the JDR threshold to half the head size as proposed by Qiu et al.~\cite{qiu_cross_2019}.

The 3D pose accuracy is measured by the Mean Per Joint Position Error (MPJPE) between estimated 3D joints $\vec{x}^j$ and ground truth 3D joints $\vec{x}^j_\text{gt}$:
$\text{MPJPE} = \frac{1}{J}\sum_{j=1}^J\lVert\vec{x}^j - \vec{x}^j_\text{gt}\rVert$.

\subsubsection{Campus and Shelf}
The Campus dataset~\cite{belagiannis_3d_2014} consists of three people interacting outdoors, captured by three calibrated cameras.
The Shelf dataset~\cite{belagiannis_3d_2014} consists of four people interacting and disassembling a shelf in a small indoor area, captured by five cameras. It is a more complex setting compared to Campus, as frequent occlusions occur between persons and with the shelf.
The same evaluation protocol as in previous works~\cite{chen_crossview_2020,dong_fast_2019,belagiannis_3d_2014,belagiannis_multiple_2015} is used, employing the 3D Percentage of Correct Parts (PCP) metric~\cite{burenius_3d_2013}. A body part is considered as correctly estimated if the average of the Euclidean distances of start and end point of the limb with the ground-truth is smaller than half the limb-length.

\subsection{Evaluation on the H3.6M Dataset}
\label{sec:eval_h36m}
\subsubsection{Implementation Details}
We adopt the network for pose estimation described in \refsec{sec:pose2d} and use two different training schemes:
\begin{inparaenum}[(i)]
\item training solely on H3.6M training data and
\item pretraining the network on person keypoints from the COCO dataset~\cite{lin_coco_2014} and finetuning on H3.6M.
\end{inparaenum}
The input resolution is set to 256$\times$256 pixels.

As is common practice in literature~\cite{qiu_cross_2019,pavlakos_harvesting_2017,tome_rethinking_2018}, we use subjects 1, 5, 6, 7, 8 for training and subjects 9 and 11 for testing. Input images are cropped using the provided ground-truth bounding box and evaluation is performed for every $5^\text{th}$ frame as subsequent frames are highly similar at the original frame rate of \SI{50}{\hertz}.

All four image streams are processed simultaneously, each on its own smart edge sensor board. The estimated 2D skeletons are transmitted to the backend, where they are triangulated, and the skeleton model is applied. We report evaluation results with and without using the proposed feedback channel. The parameters for the heatmap fusion \eqref{eqn:hm_fusion} are determined as $\alpha=0.15$ and $\beta=0.75$.

\subsubsection{Quantitative Results}
\label{sec:eval_h36m_quant}
\reftab{tab:error_2d_joint} shows evaluation results for the accuracy of the 2D pose estimation calculated on the smart edge sensors, depending on the employed feedback mode and training data. Our experiments indicate that using the feedback channel (cf. \refsec{sec:feedback}) significantly improves the JDR accuracy. The improvement is highest for the often-occluded wrist and ankle joints, \SI{5.7}{\percent} resp. \SI{5.8}{\percent} for the H3.6M-only model. For the better visible joint classes, detection is easier also without feedback and the improvement is smaller.

Pretraining the model on the COCO keypoint dataset generally improves performance, as the model trained on a larger and more varying dataset generalizes better to unknown scenes. For the stronger model, the gain from using the feedback signal is smaller, but still amounts to \SI{2}{\percent} for the wrists which are the most difficult joints to detect.

The improved 2D joint detections in turn lead to more accurate 3D poses, as becomes apparent from the results in \reftab{tab:error_3d_joint}, where 3D pose error is shown.
As in the 2D case, the improvement from the feedback channel is more significant for the weaker model and highest for ankles and wrists, around \SI{11}{\milli\meter} resp. \SI{6}{\milli\meter} for the H3.6M-only network.

In \reftab{tab:error_3d_act}, the MPJPE 3D pose error %
is shown per action category and compared to other approaches from literature. 3D pose errors after application of the resp. post-processing step or skeleton model are reported.
The recent approaches~\cite{qiu_cross_2019} and~\cite{remelli_lightweight_2020} as well as our method provide significant improvements over the older methods \cite{pavlakos_harvesting_2017} and \cite{tome_rethinking_2018}.
Comparing the models trained on H3.6M only, the results of our approach using feedback are better than~\citet{qiu_cross_2019} and~\citet{remelli_lightweight_2020} for 10 of the 15 action categories and reduce the average error. The proposed semantic feedback channel is key to this improvement over the literature.
When using additional training data, our method also achieves state-of-the-art results.

In \reftab{tab:h36m_runtime}, we compare the inference time per frame set (\ie for a set of four images) and model size of our approach with the recent approaches~\cite{qiu_cross_2019} and~\cite{remelli_lightweight_2020}. The approach of \citet{qiu_cross_2019} is an offline method with a runtime of several seconds. The approach of \citet{remelli_lightweight_2020} achieves near real-time performance on a powerful desktop GPU. The runtime of our method is still about \SI{40}{\percent} lower while running on efficient embedded sensor boards and a backend that doesn't require a GPU. Our pose estimation model, optimized for the Edge\,TPU inference accelerator, requires only \SI{12}{\mega\byte} of memory, significantly less than the models of other approaches.
\begin{table}[t]
\caption{Average inference time and model size for H3.6M dataset (Values for \citet{qiu_cross_2019} taken from~\cite{remelli_lightweight_2020}).} 
\label{tab:h36m_runtime}
\centering
\begin{threeparttable}
\begin{tabular}{l|c|c|c}
  \toprule %
  			     &\citet{qiu_cross_2019} &\citet{remelli_lightweight_2020} & Ours\\
  \midrule %
  Inference Time &  \SI{8.4}{\second} & \SI{0.040}{\second} & \textbf{\SI{0.024}{\second}}\\
  \midrule
  Model Size     &  \SI{2.1}{\giga\byte} & \SI{251}{\mega\byte} & \textbf{\SI[product-units = single]{4x12}{\mega\byte}}\\
  \bottomrule %
\end{tabular}
\end{threeparttable}
\vspace{-0.3em}
\end{table}
\begin{table}[t]
\caption{Ablation study of the impact of various components of our approach on MPJPE 3D pose error (\textit{mm}).} 
\label{tab:h36m_ablation}
\centering
\setlength{\extrarowheight}{.1em}
\begin{threeparttable}
\begin{tabular}{lccccc}
  			& \rot{MPJPE} & \rot{skeleton model} & \rot{heatmap covariance} & \rot{add. feedback} & \rot{mult. feedback}\\
  \midrule %
  w/o model, w/o fb & \multicolumn{1}{|c|}{38.08} & - & - & - & - \\
  w/o hm cov$^*$, w/o fb & \multicolumn{1}{|c|}{33.41} & \checkmark & - & - & - \\
  w/o fb            & \multicolumn{1}{|c|}{32.94} & \checkmark & \checkmark & - & - \\
  w fb (add)        & \multicolumn{1}{|c|}{30.07} & \checkmark & \checkmark & \checkmark & - \\
  w fb (mult)       & \multicolumn{1}{|c|}{30.10} & \checkmark & \checkmark & - & \checkmark \\
  w fb              & \multicolumn{1}{|c|}{\textbf{29.82}} & \checkmark & \checkmark & \checkmark & \checkmark \\
  \bottomrule %
\end{tabular}
\end{threeparttable}
\\\vspace{0.5em}
{\footnotesize $^*$skeleton model without directional heatmap covariances}
\vspace{-2em}
\end{table}

Results of an ablation study on the impacts of different parts of our proposed pipeline are shown in \reftab{tab:h36m_ablation}.
Using the skeleton model to post-process the raw 3D poses obtained by triangulation significantly improves the average MPJPE. Employing the directional covariances extracted from the heatmaps, instead of modeling uncertainties only by the confidence value, again reduces the error.
The semantic feedback further improves the result, where the proposed combination of additive and multiplicative feedback is more efficient than using only a single type.
The impact of the feedback signal for each action class can also be observed in \reftab{tab:error_3d_act}.
It improves the results for all actions for the H3.6M-only trained model, with an average improvement of \SI{3.1}{\milli\meter}. When using the stronger pose estimation model trained on additional data, the average improvement amounts to \SI{0.5}{\milli\meter}. The feedback signal is more important when the raw pose estimates are less accurate but reduces the average 3D pose error in both cases.
\begin{figure}[t]
	\vspace*{0.2em}
	\centering
	\begin{subfigure}[c]{0.24\linewidth}
		\centering
		\includegraphics[width=\linewidth]{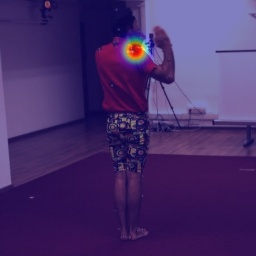}
	\end{subfigure}
	\begin{subfigure}[c]{0.24\linewidth}
		\centering
		\includegraphics[width=\linewidth]{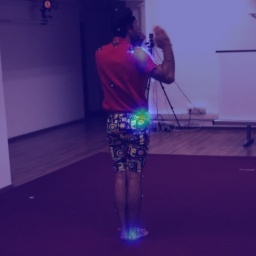}
	\end{subfigure}
	\begin{subfigure}[c]{0.24\linewidth}
		\centering
		\includegraphics[width=\linewidth]{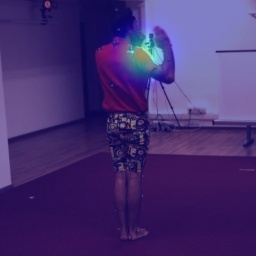}
	\end{subfigure}
	\begin{subfigure}[c]{0.24\linewidth}
		\centering
		\includegraphics[width=\linewidth]{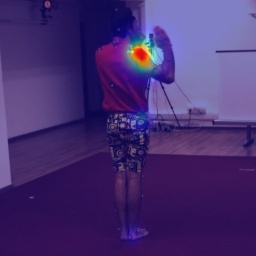}
	\end{subfigure}
	
	\vspace{0.1em}
	\begin{subfigure}[c]{0.24\linewidth}
		\centering
		\includegraphics[width=\linewidth]{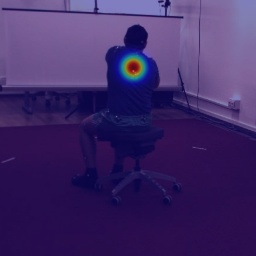}
	\end{subfigure}
	\begin{subfigure}[c]{0.24\linewidth}
		\centering
		\includegraphics[width=\linewidth]{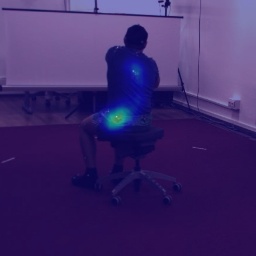}
	\end{subfigure}
	\begin{subfigure}[c]{0.24\linewidth}
		\centering
		\includegraphics[width=\linewidth]{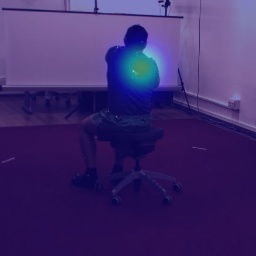}
	\end{subfigure}
	\begin{subfigure}[c]{0.24\linewidth}
		\centering
		\includegraphics[width=\linewidth]{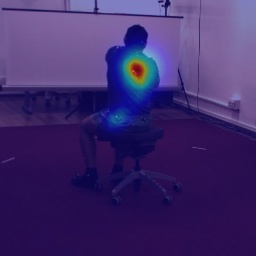}
	\end{subfigure}
	
	\vspace{0.1em}
	\begin{subfigure}[c]{0.24\linewidth}
		\centering
		\includegraphics[width=\linewidth]{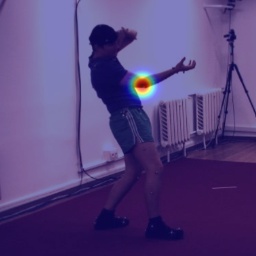}
		\caption{ground-truth}
	\end{subfigure}
	\begin{subfigure}[c]{0.24\linewidth}
		\centering
		\includegraphics[width=\linewidth]{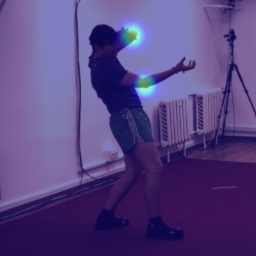}
		\caption{detected}
	\end{subfigure}
	\begin{subfigure}[c]{0.24\linewidth}
		\centering
		\includegraphics[width=\linewidth]{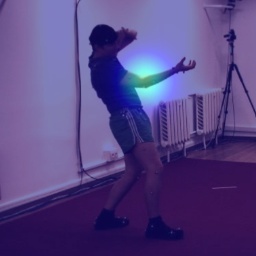}
		\caption{feedback}
	\end{subfigure}
	\begin{subfigure}[c]{0.24\linewidth}
		\centering
		\includegraphics[width=\linewidth]{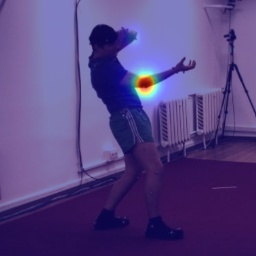}
		\caption{fused}
	\end{subfigure}
	
	\caption{Samples of the heatmap fusion approach for left wrist (Rows~1-2) and right elbow (Row~3): Detected heatmap (b) and feedback heatmap (c) are combined into the fused heatmap (d). In difficult situations such as occlusions (Rows~1-2) or left-right inversions (Row~3), using feedback results in a heatmap closer to the ground-truth (a).}
	\label{fig:feedback}
	\vspace{-1.5em}
\end{figure}

\subsubsection{Qualitative Results}
In addition, we qualitatively show how the proposed feedback loop improves the pose estimation result. \reffig{fig:feedback} shows three example situations, where the feedback heatmap helps to recover from incorrect or imprecise 2D joint detections. The images are overlaid with the respective heatmaps for a specific joint. In the first and second row, the left wrist of the actors is occluded by their body and the detected heatmap is very inaccurate. However, from the perspectives of other cameras, the joint is visible, and its 3D position can be estimated. This is reflected in the feedback heatmap which predicts the joint detection close to the ground truth location. The resulting fused heatmap, obtained by combining detection and feedback according to~\eqref{eqn:hm_fusion}, permits to accurately estimate the respective joint despite the imprecise detection. In Row~3 of \reffig{fig:feedback}, a similar situation is shown, but for the right elbow, which here cannot be distinguished from the left elbow due to the challenging pose. %

\subsection{Evaluation on the Campus and Shelf Datasets}
\label{sec:eval_campus}
\subsubsection{Implementation Details}
To process multi-person scenes, a person detector is employed together with the pose estimation model (cf. \refsec{sec:pose2d}).
The detector is trained for 130 Epochs on the person class of the COCO dataset~\cite{lin_coco_2014}, for input images of 640$\times$480~px and achieves a mAP of \SI{44.6}{\percent}.
The pose estimation network is trained for 140 epochs on COCO for person crops of 192$\times$256~px. It achieves a mAP of \SI{69.6}{\percent} in FP32-mode and \SI{68.4}{\percent} in INT8-mode on the COCO validation set using ground-truth detections.
Note, that the generic detector and pose estimation networks are employed without any fine-tuning on the evaluated datasets.

The three or five image streams of the respective dataset are processed simultaneously on the sensor boards.
The entire image is passed to the detector and image crops of the detected persons are analyzed by the pose estimation network.
To improve the processing speed, the detector is only run once per second. In between, the crops are determined based on the detections of the previous frame.
This is necessary, as alternating between models is inefficient on the Edge\,TPU, as parameter caching cannot come into effect in this case~\cite{yazdanbakhsh_edgetpu_2021}. %

On the backend, the estimated 2D poses are synchronized based on their timestamps and the framework is run in multi-person mode as detailed in \refsec{sec:multiperson}.
The feedback delay amounts to one frame during dataset processing.

\subsubsection{Quantitative Results}
\begin{table}[t]
\caption{Evaluation result on Campus and Shelf dataset: Percentage of Correct Parts (PCP) ($\%$) and average run-time of 2D and 3D pose inference. `-' means offline pre-computation.}
\label{tab:eval_campus}
\centering
\setlength{\tabcolsep}{3.3pt}
\begin{threeparttable}
\begin{tabular}{l|cccc|cc}
  \toprule
  													    &\multicolumn{4}{c|}{PCP ($\%$)}&\multicolumn{2}{c}{Inference Time}\\
  \cmidrule{2-7}
   Campus 												& Actor1 & Actor2 & Actor3 & Avg & 2D pose & 3D pose\\
  \midrule
  Belagiannis et al.~\cite{belagiannis_multiple_2015} & 83.0 & 73.0 & 78.0 & 78.0 & -  & 1\,s\\
  Dong et al.~\cite{dong_fast_2019}					   & 97.6 & 93.3 & 98.0 & 96.3 & -  & 105\,ms\\
  Chen et al.~\cite{chen_crossview_2020}			   & 97.1 & \textbf{94.1} & \textbf{98.6} & 96.6 & - & \textbf{1.6\,ms}\\
  Ours, w/o fb										   & 98.8 & 93.4 & 97.5 & 96.6 & \textbf{30\,ms} & 8.8\,ms\\
  Ours, w fb 										   & \textbf{99.2} & 93.6 & 98.3 & \textbf{97.0} & \textbf{30\,ms} & 8.8\,ms\\
  \bottomrule
  \toprule
   Shelf 											   & Actor1 & Actor2 & Actor3 & Avg & 2D pose & 3D pose\\
  \midrule
  Belagiannis et al.~\cite{belagiannis_multiple_2015} & 75.0 & 67.0 & 86.0 & 76.0 & -  & 1\,s\\
  Dong et al.~\cite{dong_fast_2019}					   & 98.8 & 94.1 & \textbf{97.8} & 96.9 & - & 105\,ms\\
  Chen et al.~\cite{chen_crossview_2020}			   & \textbf{99.6} & 93.2 & 97.5 & 96.8 & - & \textbf{3.1\,ms}\\
  Ours, w/o fb										   & 99.4 & 94.6 & 96.8 & 96.9 & \textbf{40\,ms} & 20\,ms\\
  Ours, w fb 										   & 99.3 & \textbf{95.7} & 97.3 & \textbf{97.4} & \textbf{40\,ms} & 20\,ms\\
  \bottomrule
\end{tabular}
\end{threeparttable}
\vspace{-1.em}
\end{table}
We report evaluation results of PCP score and runtime on the Campus and Shelf datasets in \reftab{tab:eval_campus}. and compare our method with other approaches: Belagiannis et al.~\cite{belagiannis_multiple_2015} were among the first to propose 3D PSM-based multi-person pose estimation and exploit temporal consistency in videos. Dong et al.~\cite{dong_fast_2019} propose to reduce the PSM state-space %
and exploit appearance information for data association. Chen et al.~\cite{chen_crossview_2020} propose a fast iterative triangulation scheme performing data association in 3D space.

In terms of PCP score, our method largely outperforms the older method~\cite{belagiannis_multiple_2015} and is on par with the recent approaches~\cite{dong_fast_2019,chen_crossview_2020}. The overall result is improved by our method using feed\-back for both Campus and Shelf dataset in comparison to the literature. The improvement is most significant for Actor2 of the Shelf dataset, whose arms are often severely occluded, which can be resolved by the semantic feedback signal.

In terms of processing speed, our method does not reach the high frame rates of Chen et al.~\cite{chen_crossview_2020} but achieves significant improvements over~\cite{dong_fast_2019} and~\cite{belagiannis_multiple_2015}.
Furthermore, 2D poses are estimated online, at run-times of \SIrange{30}{40}{\milli\second} per frame, while other methods use offline pre-computed keypoint detections.
Our method is the only approach in the comparison providing a fully online multi-person pose estimation.
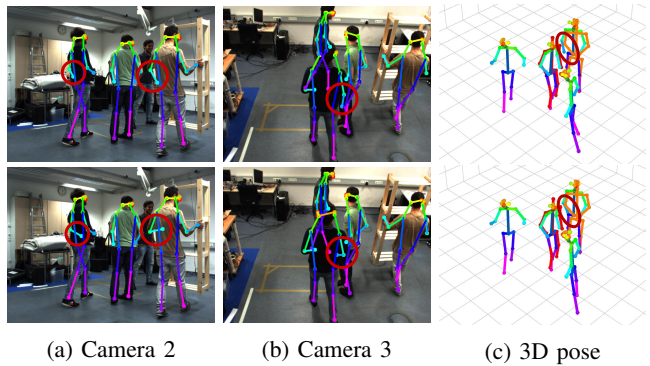
\begin{figure}[t]
\vspace{0.2em}
	\centering
	\begin{subfigure}[c]{0.31\linewidth}
		\centering
		\begin{tikzpicture}
			\definecolor{red}{rgb}{0.8,0.0,0.0}
			\node[inner sep=0] (image1) at (0, 0) {\includegraphics[width=\linewidth]{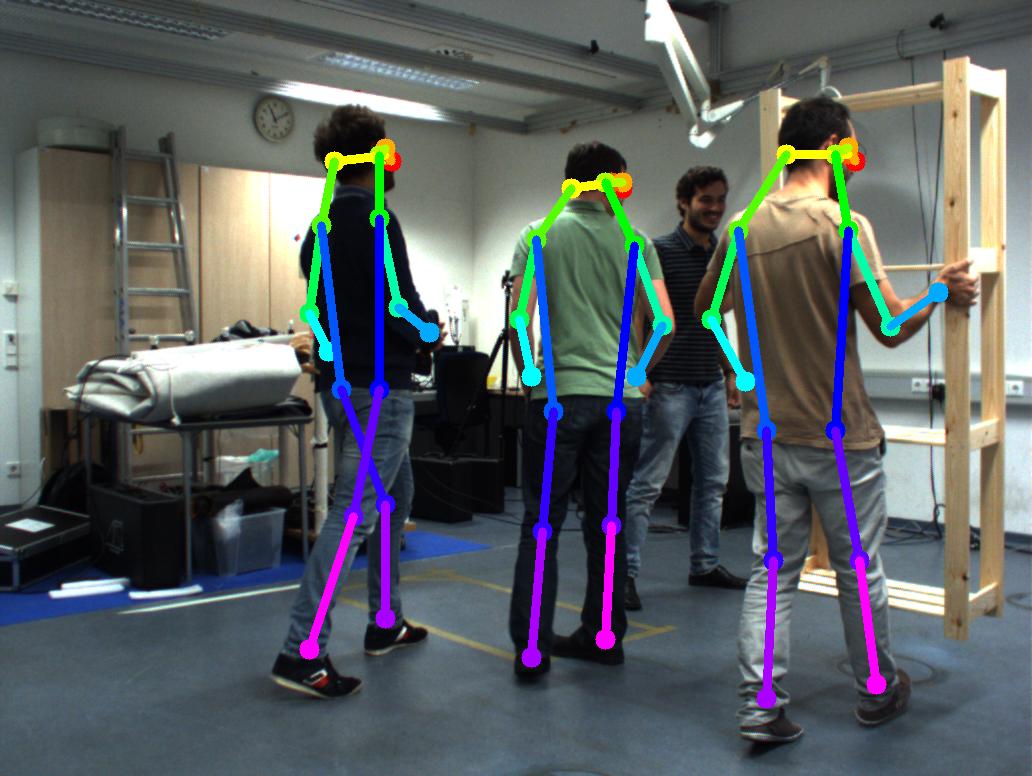}};
			\draw[line width=0.4mm, red] (0.55, 0.1) circle (2mm and 2mm);
			\draw[line width=0.4mm, red] (-0.5, 0.15) circle (1.5mm and 1.5mm);
		\end{tikzpicture}
	\end{subfigure}
	\begin{subfigure}[c]{0.31\linewidth}
		\centering
		\begin{tikzpicture}
			\definecolor{red}{rgb}{0.8,0.0,0.0}
			\node[inner sep=0] (image1) at (0, 0) {\includegraphics[width=\linewidth]{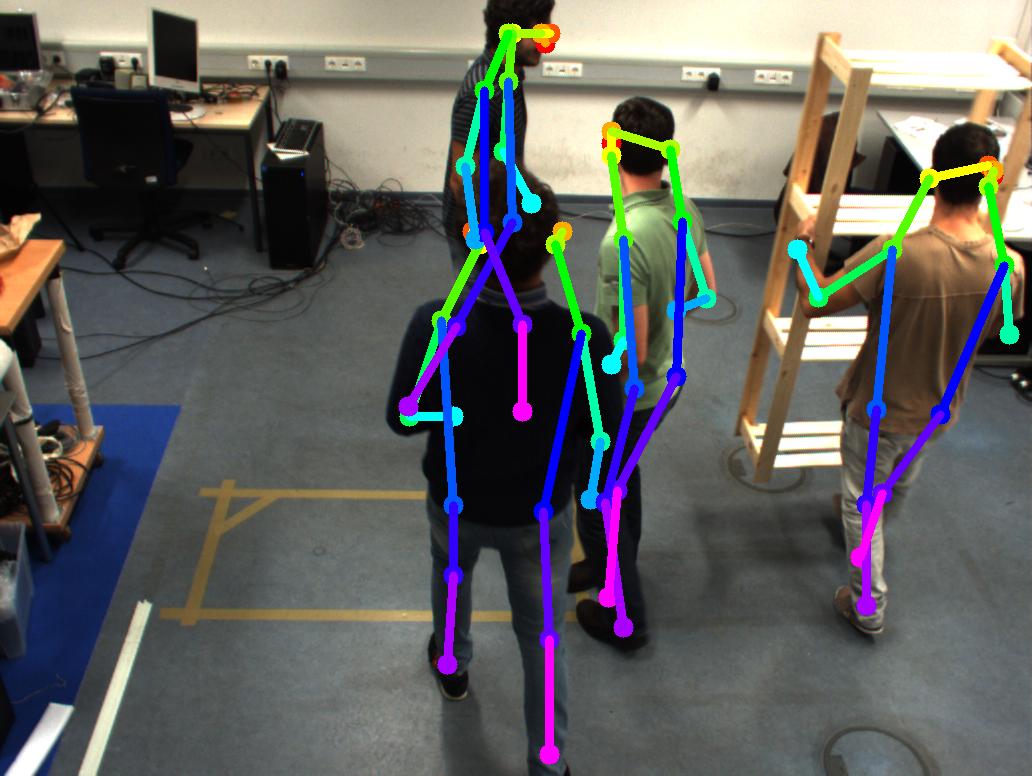}};
			\draw[line width=0.4mm, red] (0.2, -0.2) circle (2mm and 2mm);
		\end{tikzpicture}
	\end{subfigure}
	\begin{subfigure}[c]{0.31\linewidth}
		\centering
		\begin{tikzpicture}
			\definecolor{red}{rgb}{0.7,0.0,0.0}
			\node[inner sep=0] (image1) at (0, 0) {\includegraphics[trim=40mm 15mm 90mm 75mm,clip,width=\linewidth]{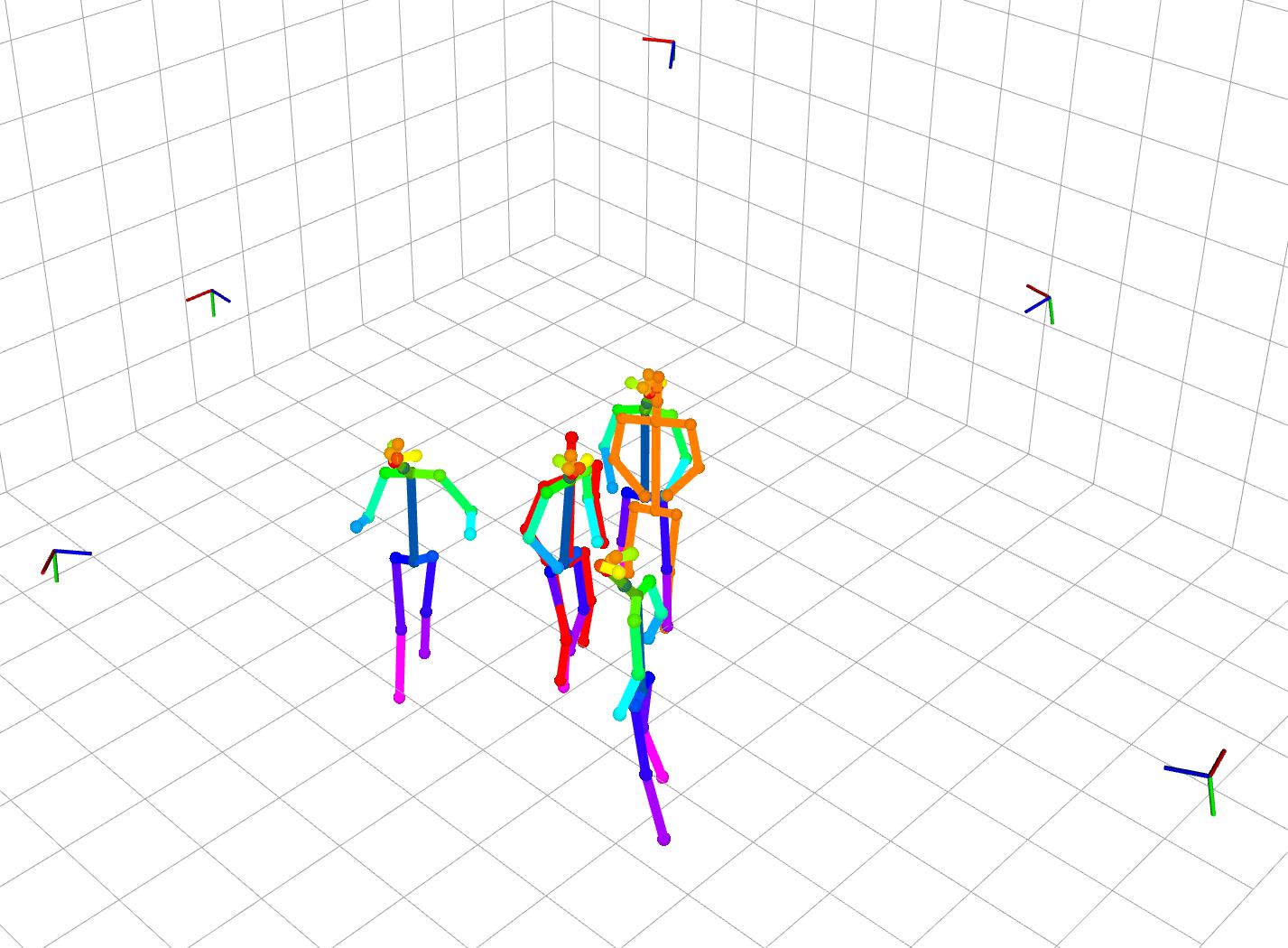}};
			\draw[line width=0.4mm, rotate around={25:(0.33, 0.47)}, red] (0.33, 0.47) circle (1.2mm and 2.2mm);
		\end{tikzpicture}
	\end{subfigure}
	
	\vspace{0.1em}
	\begin{subfigure}[c]{0.31\linewidth}
		\centering
		\begin{tikzpicture}
			\definecolor{red}{rgb}{0.8,0.0,0.0}
			\node[inner sep=0] (image1) at (0, 0) {\includegraphics[width=\linewidth]{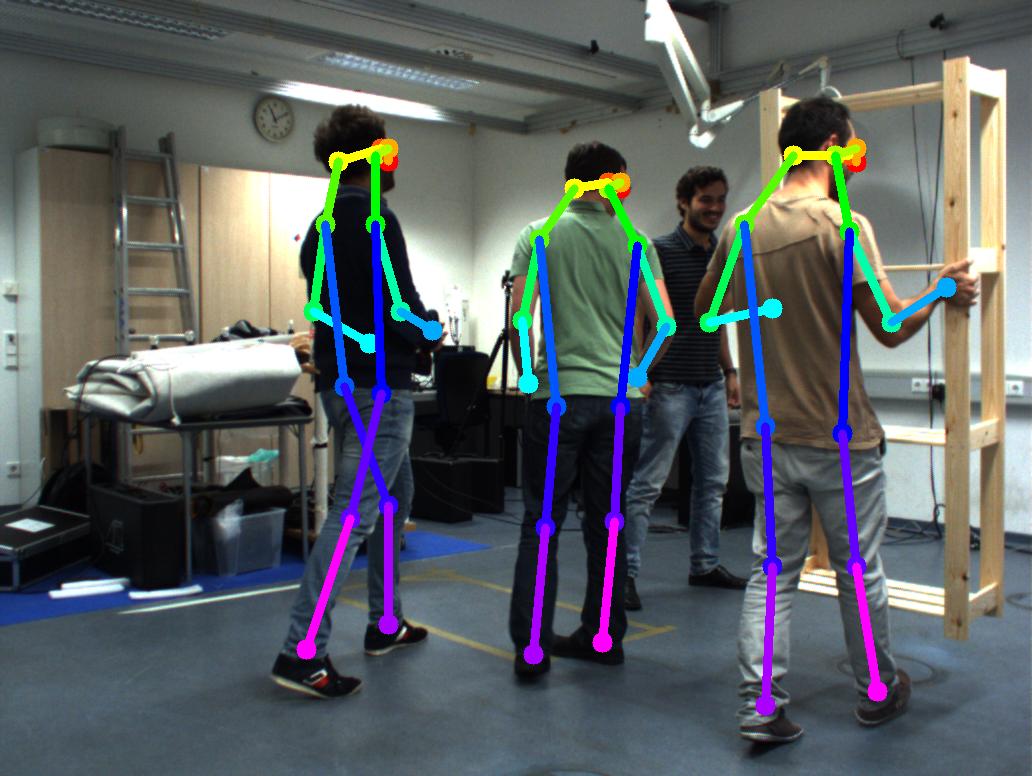}};
			\draw[line width=0.4mm, red] (0.6, 0.2) circle (2mm and 2mm);
			\draw[line width=0.4mm, red] (-0.45, 0.15) circle (1.5mm and 1.5mm);
		\end{tikzpicture}
		\caption{Camera~2}
	\end{subfigure}
	\begin{subfigure}[c]{0.31\linewidth}
		\centering
		\begin{tikzpicture}
			\definecolor{red}{rgb}{0.8,0.0,0.0}
			\node[inner sep=0] (image1) at (0, 0) {\includegraphics[width=\linewidth]{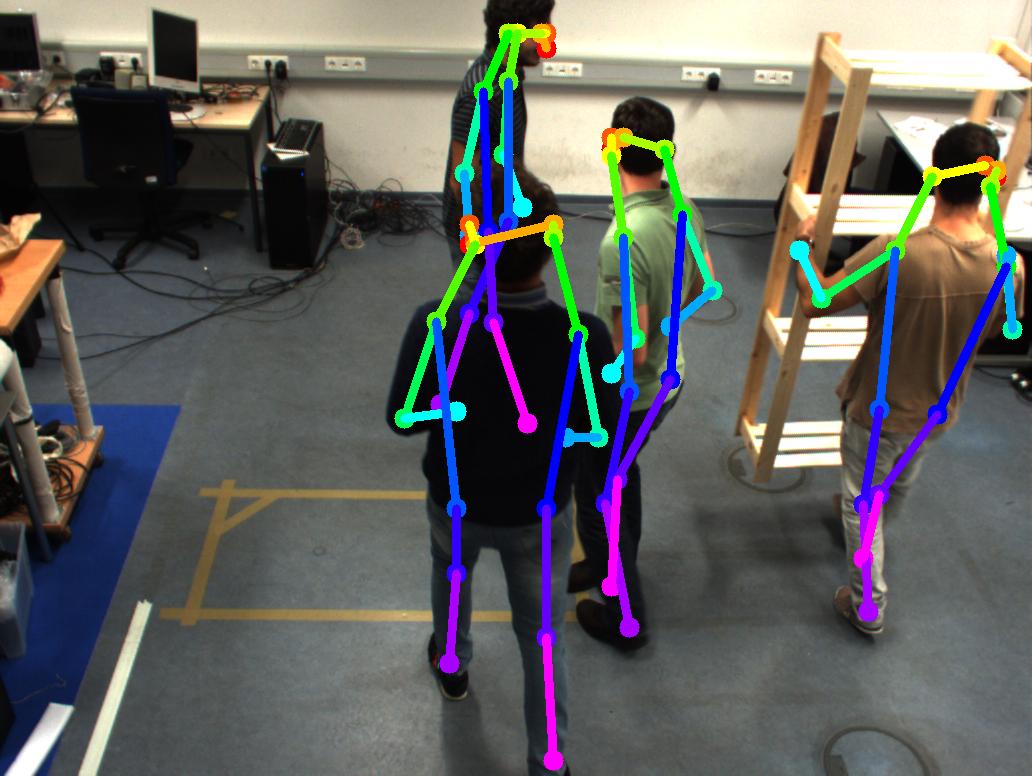}};
			\draw[line width=0.4mm, red] (0.2, -0.1) circle (2mm and 2mm);
		\end{tikzpicture}
		\caption{Camera~3}
	\end{subfigure}
	\begin{subfigure}[c]{0.31\linewidth}
		\centering
		\begin{tikzpicture}
			\definecolor{red}{rgb}{0.7,0.0,0.0}
			\node[inner sep=0] (image1) at (0, 0) {\includegraphics[trim=40mm 15mm 90mm 75mm,clip,width=\linewidth]{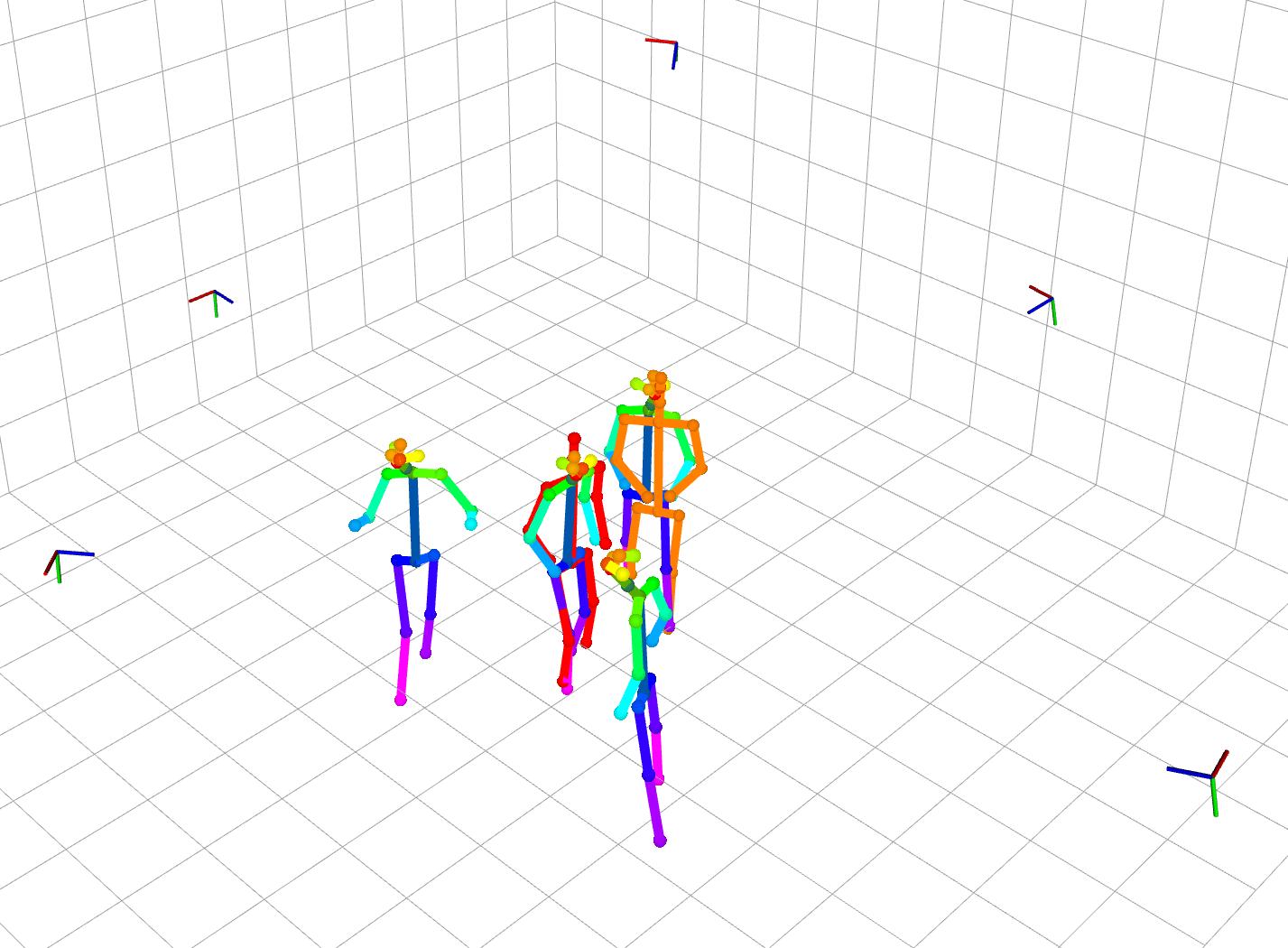}};
			\draw[line width=0.4mm, rotate around={25:(0.35, 0.45)}, red] (0.35, 0.45) circle (1.2mm and 2.2mm);
		\end{tikzpicture}
		\caption{3D pose}
	\end{subfigure}
	\caption{Evaluation on Shelf dataset: 2D pose detections and estimated 3D pose without (top) and with feedback (bottom). 3D annotations for Actor1 (red) and Actor2 (orange). Highlighted are improvements due to the feedback signal.}
	\label{fig:eval_shelf}
\end{figure}
\subsubsection{Qualitative Results}
\reffig{fig:eval_shelf} shows an exemplary scene of the Shelf dataset. The proposed semantic feedback improves the estimation of occluded wrist joints in 2D and 3D. Annotations for evaluation are only provided for two of the four actors in this scene.

\subsection{Experiments in Multi-Person Scenes}
\label{sec:eval_realworld}
We further evaluate the proposed framework in online ex\-per\-i\-ments in multi-person scenarios in our lab.
\subsubsection{Implementation Details}
16 sensor boards are mounted under the ceiling of our lab in a roughly 12$\times$16~m area.
The cameras face downwards towards the center and run at \SI{30}{\hertz} and VGA resolution.
We conduct experiments with a subset of 4 cameras, similar to the setting of the H3.6M dataset, as well as with all 16 sensors to demonstrate the scalability of our method to large-scale camera systems.
The same detection and pose estimation models as for the Campus and Shelf dataset are employed in the experiments and the pipeline runs in multi-person mode (cf. \refsec{sec:multiperson}).
\subsubsection{Quantitative Results}
To analyze the consistency of the online pose estimation, we evaluate the error between detected 2D poses and fused 3D poses reprojected into the camera views in \reftab{tab:error_2d_reproj}. The reprojection error decreases for all joints when using semantic feedback, indicating that the locally estimated 2D poses are more consistent with the globally fused 3D poses through the proposed feedback architecture. The error is slightly higher with 16 than with 4 sensors, probably due to the more difficult camera calibration and synchronization in the large-scale setup.
\begin{table}
\caption{Evaluation in own experiments with up to 16 cameras and 6 persons: Reprojection error (px) per joint class between detected 2D poses and fused 3D poses.}
\label{tab:error_2d_reproj}
\centering
\setlength{\tabcolsep}{3pt}
\begin{threeparttable}
\begin{tabular}{lcc|ccccccc}
  \toprule
   Feedback & Cams & Pers & Hips & Knees & Ankls & Shlds & Elbs & Wrists & \textbf{Avg}\\
  \midrule
   w/o fb   & 4  & 1-4 & 5.4 & 4.6  & 5.0  & 2.8  & 4.0 & 5.2 & 4.2 \\
   w fb     & 4  & 1-4 & 4.4  & \textbf{3.5}  & \textbf{3.4}  & \textbf{2.3}  & \textbf{3.2} & \textbf{3.7} & \textbf{3.3} \\ %
   w/o fb   & 16 & 6 & 5.4  & 5.2  & 6.4  & 3.9  & 5.1 & 6.4 & 5.1 \\
   w fb     & 16 & 6 & \textbf{4.3} & 3.8 & 4.7 & 3.4 & 4.0 & 4.9 & 4.1 \\ %
   \bottomrule
\end{tabular}
\end{threeparttable}
\vspace{-1.em}
\end{table}
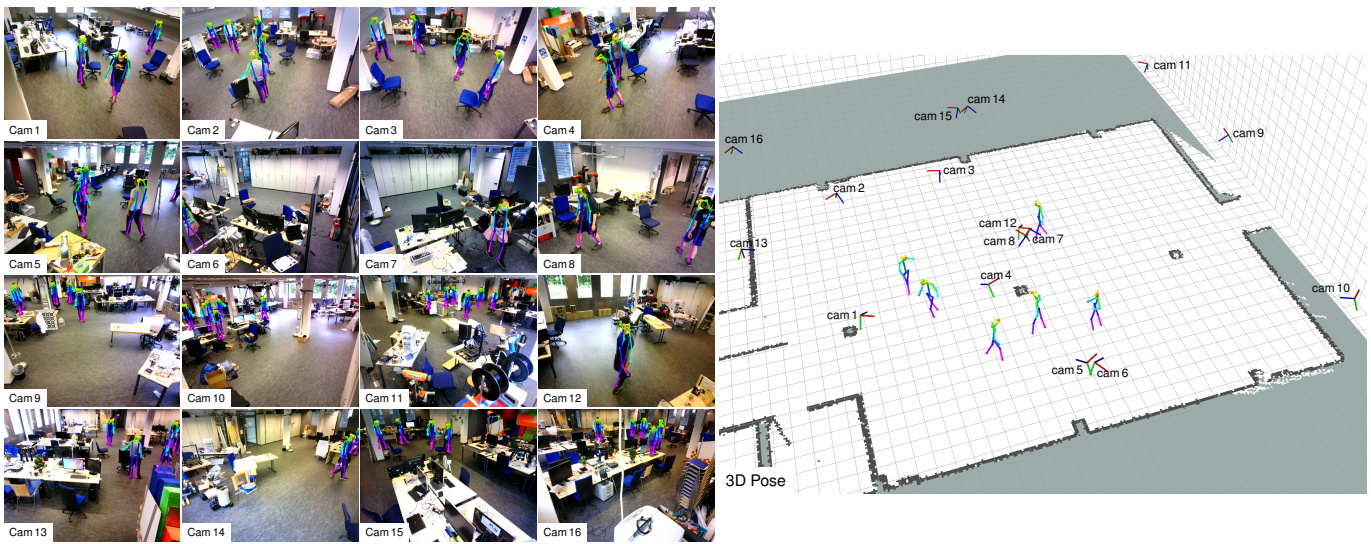
\begin{figure*}[!ht]
	\centering
	\resizebox{1.0\linewidth}{!}{%
\begin{tikzpicture}
    \begin{scope}[scale=0.8, transform shape] %
        \node[inner sep=0,anchor=north west] (image1) at (0,0) {\includegraphics{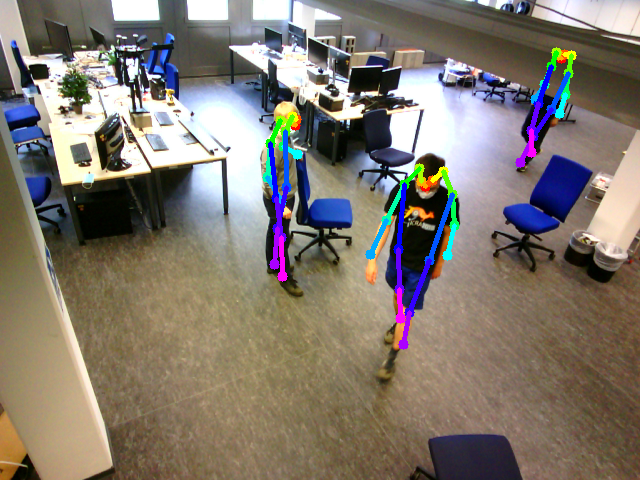}};
        \node[inner sep=0,anchor=north west,xshift=0.2cm] (image2) at (image1.north east) {\includegraphics{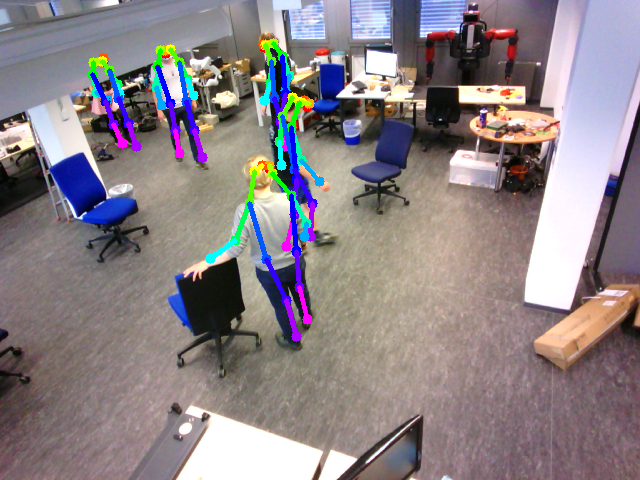}};
        \node[inner sep=0,anchor=north west,xshift=0.2cm] (image3) at (image2.north east) {\includegraphics{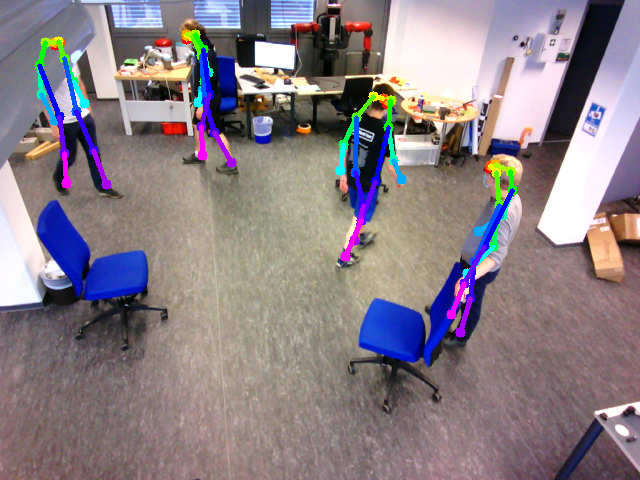}};
        \node[inner sep=0,anchor=north west,xshift=0.2cm] (image4) at (image3.north east) {\includegraphics{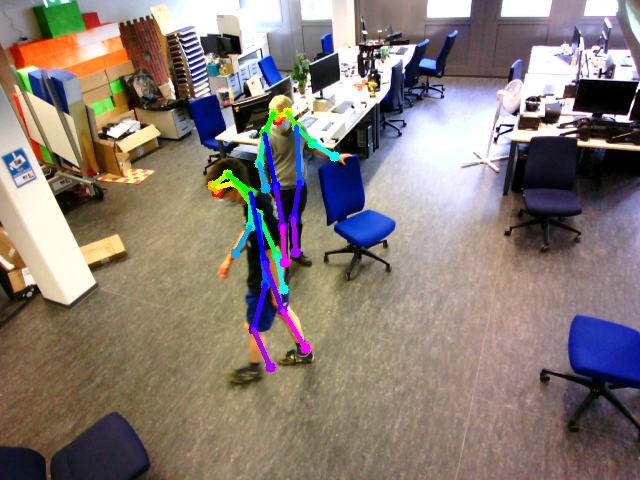}};
        
        \node[inner sep=0,anchor=north west,yshift=-0.2cm] (image5) at (image1.south west) {\includegraphics{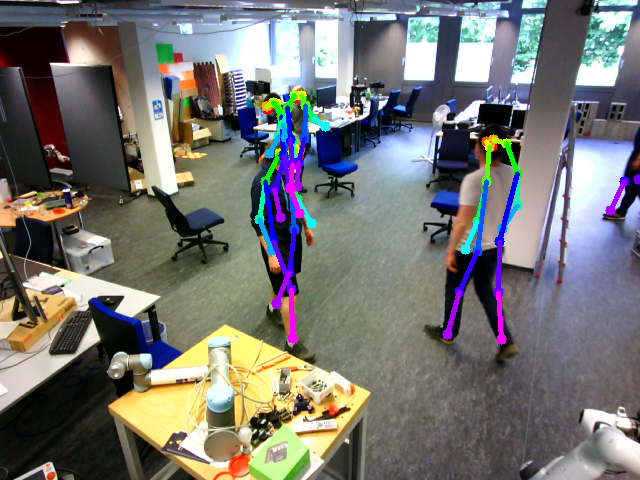}};
        \node[inner sep=0,anchor=north west,xshift=0.2cm] (image6) at (image5.north east) {\includegraphics{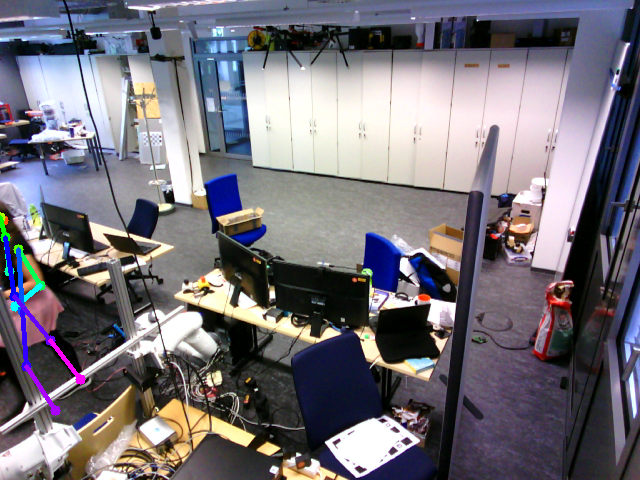}};
        \node[inner sep=0,anchor=north west,xshift=0.2cm] (image7) at (image6.north east) {\includegraphics{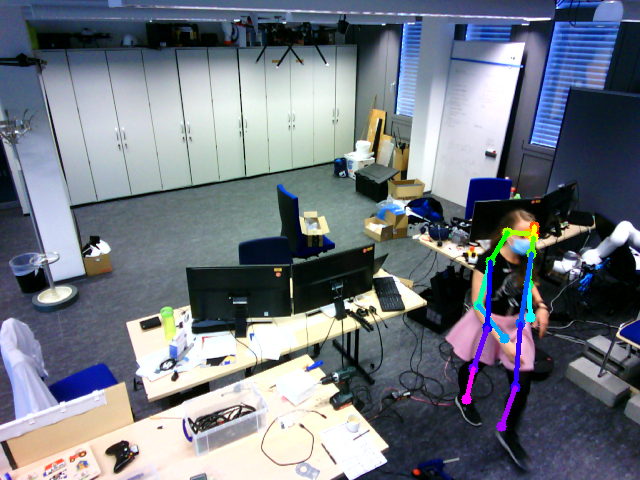}};
        \node[inner sep=0,anchor=north west,xshift=0.2cm] (image8) at (image7.north east) {\includegraphics{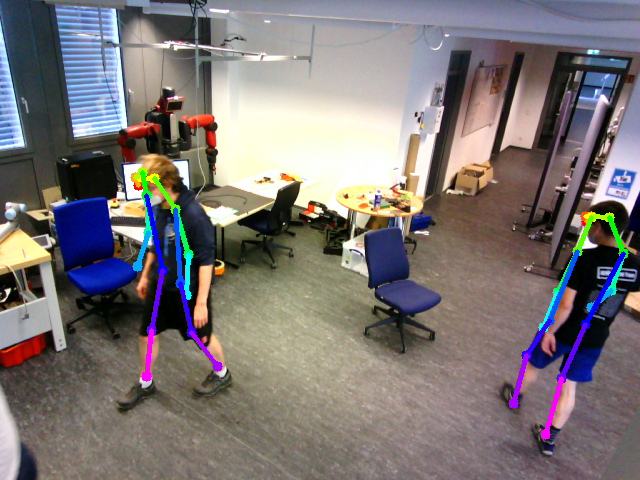}};
        
        \node[inner sep=0,anchor=north west,yshift=-0.2cm] (image9) at (image5.south west) {\includegraphics{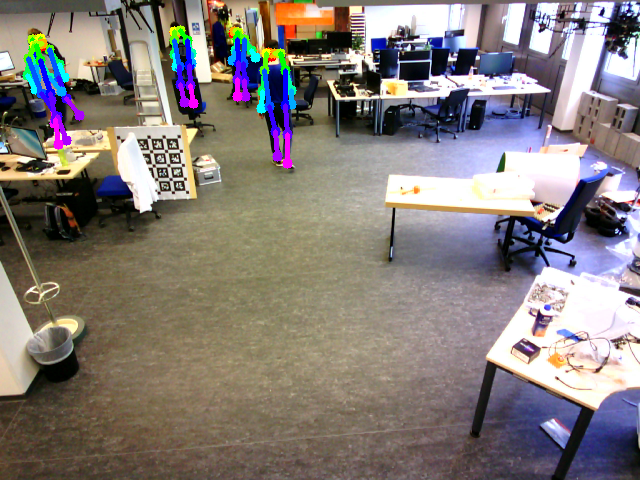}};
        \node[inner sep=0,anchor=north west,xshift=0.2cm] (image10) at (image9.north east) {\includegraphics{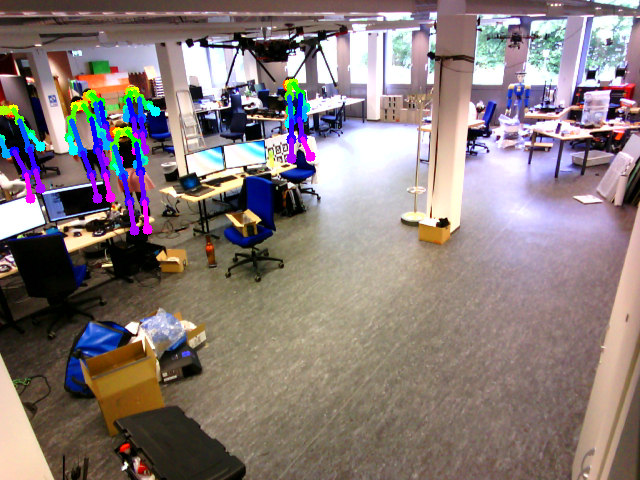}};
        \node[inner sep=0,anchor=north west,xshift=0.2cm] (image11) at (image10.north east) {\includegraphics{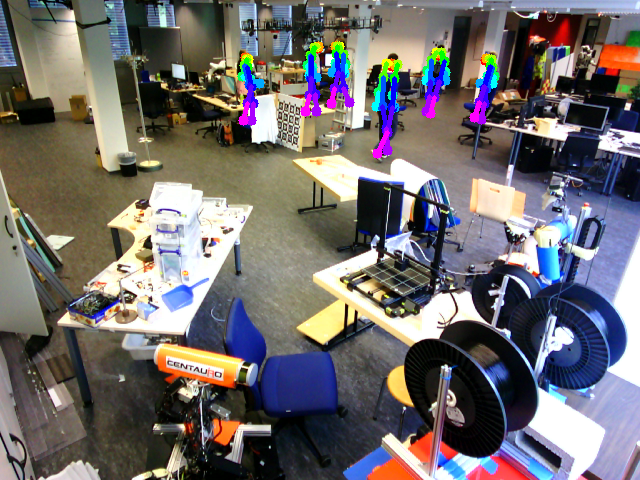}};
        \node[inner sep=0,anchor=north west,xshift=0.2cm] (image12) at (image11.north east) {\includegraphics{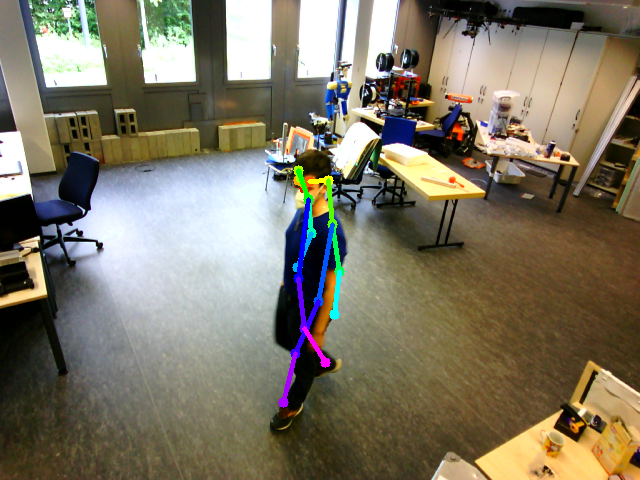}};
        
        \node[inner sep=0,anchor=north west,yshift=-0.2cm] (image13) at (image9.south west) {\includegraphics{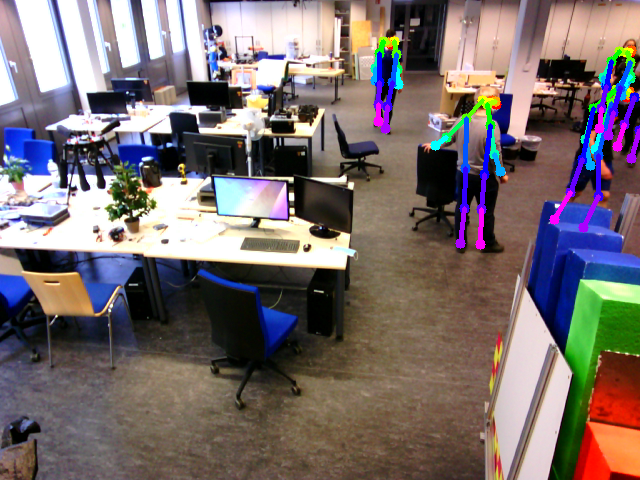}};
        \node[inner sep=0,anchor=north west,xshift=0.2cm] (image14) at (image13.north east) {\includegraphics{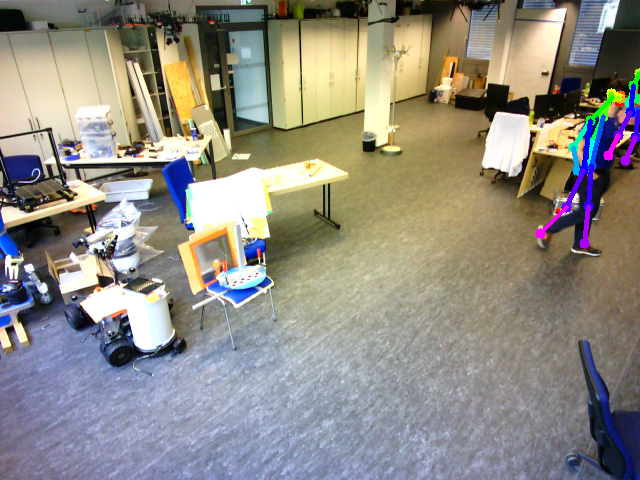}};
        \node[inner sep=0,anchor=north west,xshift=0.2cm] (image15) at (image14.north east) {\includegraphics{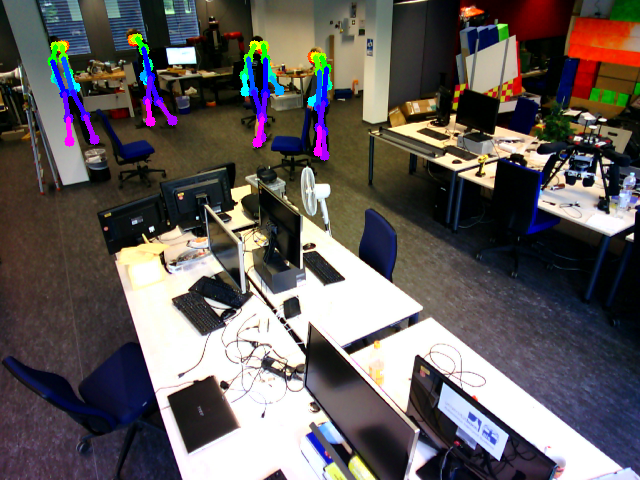}};
        \node[inner sep=0,anchor=north west,xshift=0.2cm] (image16) at (image15.north east) {\includegraphics{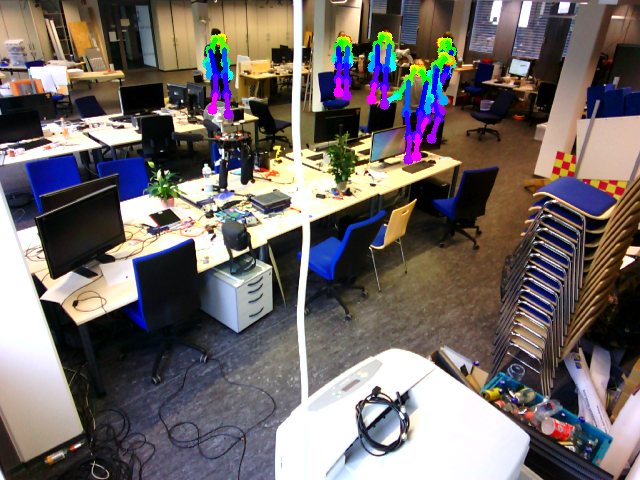}};
        
        \node[label,scale=3.5, anchor=south west, xshift=-0.2mm, yshift=-0.2mm, rectangle, fill=white, align=center, font=\scriptsize\sffamily] (n_1) at (image1.south west) {Cam\,1};
        \node[label,scale=3.5, anchor=south west, xshift=-0.2mm, yshift=-0.2mm, rectangle, fill=white, align=center, font=\scriptsize\sffamily] (n_2) at (image2.south west) {Cam\,2};
        \node[label,scale=3.5, anchor=south west, xshift=-0.2mm, yshift=-0.2mm, rectangle, fill=white, align=center, font=\scriptsize\sffamily] (n_3) at (image3.south west) {Cam\,3};
        \node[label,scale=3.5, anchor=south west, xshift=-0.2mm, yshift=-0.2mm, rectangle, fill=white, align=center, font=\scriptsize\sffamily] (n_4) at (image4.south west) {Cam\,4};
        \node[label,scale=3.5, anchor=south west, xshift=-0.2mm, yshift=-0.2mm, rectangle, fill=white, align=center, font=\scriptsize\sffamily] (n_5) at (image5.south west) {Cam\,5};
        \node[label,scale=3.5, anchor=south west, xshift=-0.2mm, yshift=-0.2mm, rectangle, fill=white, align=center, font=\scriptsize\sffamily] (n_6) at (image6.south west) {Cam\,6};
        \node[label,scale=3.5, anchor=south west, xshift=-0.2mm, yshift=-0.2mm, rectangle, fill=white, align=center, font=\scriptsize\sffamily] (n_7) at (image7.south west) {Cam\,7};
        \node[label,scale=3.5, anchor=south west, xshift=-0.2mm, yshift=-0.2mm, rectangle, fill=white, align=center, font=\scriptsize\sffamily] (n_8) at (image8.south west) {Cam\,8};
        \node[label,scale=3.5, anchor=south west, xshift=-0.2mm, yshift=-0.2mm, rectangle, fill=white, align=center, font=\scriptsize\sffamily] (n_9) at (image9.south west) {Cam\,9};
        \node[label,scale=3.5, anchor=south west, xshift=-0.2mm, yshift=-0.2mm, rectangle, fill=white, align=center, font=\scriptsize\sffamily] (n_10) at (image10.south west) {Cam\,10};
        \node[label,scale=3.5, anchor=south west, xshift=-0.2mm, yshift=-0.2mm, rectangle, fill=white, align=center, font=\scriptsize\sffamily] (n_11) at (image11.south west) {Cam\,11};
        \node[label,scale=3.5, anchor=south west, xshift=-0.2mm, yshift=-0.2mm, rectangle, fill=white, align=center, font=\scriptsize\sffamily] (n_12) at (image12.south west) {Cam\,12};
        \node[label,scale=3.5, anchor=south west, xshift=-0.2mm, yshift=-0.2mm, rectangle, fill=white, align=center, font=\scriptsize\sffamily] (n_13) at (image13.south west) {Cam\,13};
        \node[label,scale=3.5, anchor=south west, xshift=-0.2mm, yshift=-0.2mm, rectangle, fill=white, align=center, font=\scriptsize\sffamily] (n_14) at (image14.south west) {Cam\,14};
        \node[label,scale=3.5, anchor=south west, xshift=-0.2mm, yshift=-0.2mm, rectangle, fill=white, align=center, font=\scriptsize\sffamily] (n_15) at (image15.south west) {Cam\,15};
        \node[label,scale=3.5, anchor=south west, xshift=-0.2mm, yshift=-0.2mm, rectangle, fill=white, align=center, font=\scriptsize\sffamily] (n_16) at (image16.south west) {Cam\,16};
    \end{scope}
    
    \node[inner sep=0,anchor=north west,xshift=0.3cm,yshift=-3cm] (image3D) at (image4.north east) {\includegraphics[trim=20mm 0mm 20mm 0mm,clip]{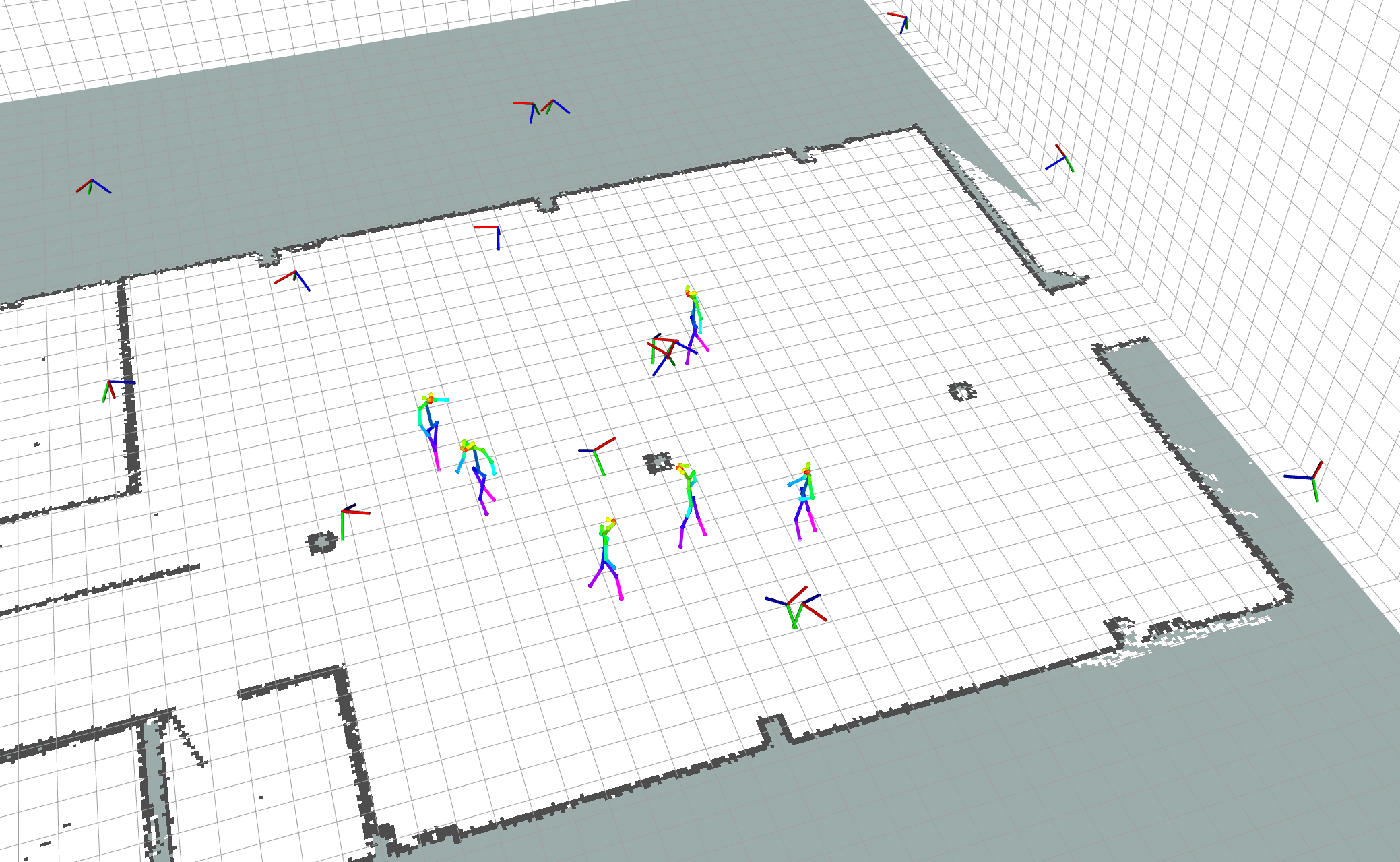}};
    \node[label,scale=4., anchor=south west, xshift=-0.2mm, yshift=-0.2mm, rectangle, fill=white, align=center, font=\scriptsize\sffamily] (n_3D) at (image3D.south west) {3D Pose};
    \node[label, anchor=south west, xshift=6.3cm, yshift=-16.7cm, rectangle, align=center, font=\scriptsize\sffamily, scale=3] (l_cam1) at (image3D.north west) {cam\,1};
    \node[label, anchor=south west, xshift=6.7cm, yshift=-8.8cm, rectangle, align=center, font=\scriptsize\sffamily, scale=3] (l_cam2) at (image3D.north west) {cam\,2};
    \node[label, anchor=south west, xshift=13.5cm, yshift=-7.7cm, rectangle, align=center, font=\scriptsize\sffamily, scale=3] (l_cam3) at (image3D.north west) {cam\,3};
    \node[label, anchor=south west, xshift=15.8cm, yshift=-14.2cm, rectangle, align=center, font=\scriptsize\sffamily, scale=3] (l_cam4) at (image3D.north west) {cam\,4};
    \node[label, anchor=south west, xshift=20.2cm, yshift=-20.1cm, rectangle, align=center, font=\scriptsize\sffamily, scale=3] (l_cam5) at (image3D.north west) {cam\,5};
    \node[label, anchor=south west, xshift=23.cm, yshift=-20.3cm, rectangle, align=center, font=\scriptsize\sffamily, scale=3] (l_cam6) at (image3D.north west) {cam\,6};
    \node[label, anchor=south west, xshift=19.cm, yshift=-12.cm, rectangle, align=center, font=\scriptsize\sffamily, scale=3] (l_cam7) at (image3D.north west) {cam\,7};
    \node[label, anchor=south west, xshift=16cm, yshift=-12.1cm, rectangle, align=center, font=\scriptsize\sffamily, scale=3] (l_cam8) at (image3D.north west) {cam\,8};
    \node[label, anchor=south west, xshift=31.4cm, yshift=-5.4cm, rectangle, align=center, font=\scriptsize\sffamily, scale=3] (l_cam9) at (image3D.north west) {cam\,9};
    \node[label, anchor=south west, xshift=36.3cm, yshift=-15.cm, rectangle, align=center, font=\scriptsize\sffamily, scale=3] (l_cam10) at (image3D.north west) {cam\,10};
    \node[label, anchor=south west, xshift=26.5cm, yshift=-1.2cm, rectangle, align=center, font=\scriptsize\sffamily, scale=3] (l_cam11) at (image3D.north west) {cam\,11};
    \node[label, anchor=south west, xshift=15.7cm, yshift=-11.cm, rectangle, align=center, font=\scriptsize\sffamily, scale=3] (l_cam12) at (image3D.north west) {cam\,12};
    
    \node[label, anchor=south west, xshift=0.3cm, yshift=-12.2cm, rectangle, align=center, font=\scriptsize\sffamily, scale=3] (l_cam13) at (image3D.north west) {cam\,13};
    
    \node[label, anchor=south west, xshift=15.cm, yshift=-3.3cm, rectangle, align=center, font=\scriptsize\sffamily, scale=3] (l_cam14) at (image3D.north west) {cam\,14};
    \node[label, anchor=south west, xshift=11.7cm, yshift=-4.4cm, rectangle, align=center, font=\scriptsize\sffamily, scale=3] (l_cam15) at (image3D.north west) {cam\,15};
    
    \node[label, anchor=south west, xshift=0.cm, yshift=-5.8cm, rectangle, align=center, font=\scriptsize\sffamily, scale=3] (l_cam16) at (image3D.north west) {cam\,16};
\end{tikzpicture}
}
	\caption{Evaluation in multi-person scenarios: Estimated 3D poses reprojected into the corresponding camera image.}
	\label{fig:realworld}
	\vspace{-1.em}
\end{figure*}
\subsubsection{Qualitative Results}
An exemplary real-world scene from the experiments conducted in our lab is shown in \reffig{fig:realworld}. The 16 camera views contain up to six persons and complex, cluttered backgrounds.
Estimated 3D poses are reprojected onto the corresponding images to provide a visual evaluation. The reprojected skeletons closely fit the persons in the images, indicating that 3D and 2D poses are reliably estimated. People are reliably detected also at large distances to the cameras and occlusions by objects or other people can be resolved through the multi-view architecture and the semantic feedback.

The human poses are estimated online, in real time, and could directly be used, \eg for human-robot interaction. %
Note, that the camera images are not transmitted during operation of our framework but are only shown for visualization.
A video of the experiments is available on our website\footnote{\href{https://www.ais.uni-bonn.de/videos/RSS_2021_Bultmann}{https://www.ais.uni-bonn.de/videos/RSS\_2021\_Bultmann}}.

\subsubsection{Run-time Analysis}
\label{sec:runtime}
The average processing time per image crop on the sensor boards consists of \SI{4.5}{\milli\second} for pose estimation on the TPU and \SI{6}{\milli\second} on the ARM-CPU for pre- and post-processing and sums to \SI{10.5}{\milli\second} per detected person.
Once per second, the person detector requires additional \SI{20}{\milli\second} on the TPU.
Up to three persons can thus be tracked at the full camera frame rate of \SI{30}{\hertz}, six persons still at \SI{15}{\hertz}.

The backend processing on a desktop PC with an Intel i9-9900K CPU takes \SI{10.7}{\milli\second} in average per frame set for the 4-camera setup and \SI{60.8}{\milli\second} during the experiments with 16 cameras and six persons. Especially the computational load of multi-view triangulation grows with larger number of cameras.

Camera images and semantic feedback are processed asynchronously on the sensors during the online experiments, the frequencies of the feedback and feed-forward parts of the pipeline do not need to be balanced. The most recent feedback message not older than a threshold is used for a camera image.
The average pipeline delay $\Delta t$ including processing on sensors and backend as well as network and synchronization delays sums to \SI{89}{\milli\second} in the 4-camera setup and to \SI{200}{\milli\second} with 16 cameras.
This delay does not limit the feed-forward frequency of pose inference due to the asynchronous parallel processing. The latency is compensated by the prediction step in the feedback channel (cf. \refsec{sec:feedback}).

\subsubsection{Network Bandwidth and Power Consumption}
\label{sec:network}
The network usage when processing a \SI{30}{\hertz} video stream only amounts to \SI{15}{\kilo\byte\per\second} per detected person, as only semantic skeletons are transmitted between sensors and backend. This is an over \SI{99}{\percent} reduction of bandwidth compared to \SI{27}{\mega\byte\per\second} when transmitting the raw VGA images. The power consumption of a sensor board was measured as approx. \SI{7}{\watt} when running inference on the \SI{30}{\hertz} multi-person video stream.
 
\section{Conclusion and Future Work}
\label{sec:Conclusion}
In this work, we proposed a novel method for real-time 3D human pose estimation using a network of smart edge sensors.
Our main idea is to process each camera view on-device, transmit only semantic information to a backend where it is fused into a 3D skeleton, and implement a 3D\,/\,2D semantic feedback channel which lets the local semantic models incorporate fused multi-view information.
The pipeline is able to track up to three persons at \SI{30}{\hertz} and up to six persons at \SI{15}{\hertz}, achieving real-time performance. It is evaluated on the H3.6M, Campus, and Shelf datasets where it is shown to achieve state-of-the-art results, as well as on own data in scenarios with up to 16 cameras and six persons.

In future research, we plan to use the estimated human pose information to enable safe human-robot interaction and anticipative robot behavior in a workspace shared with people. Mobile robots carrying a smart sensor board can participate in the network for collaborative perception and add further viewpoints. The semantic scene model could be extended to also include objects and scene geometry.
Furthermore, using a more elaborate motion model in the prediction step could compensate better for the pipeline delay and improve the feedback signal, especially for fast motions.
 
\section*{Acknowledgments}
This work was funded by grant BE 2556/16-2 of the German Research Foundation (DFG), a Google faculty research award, and Fraunhofer IAIS.

\bibliographystyle{plainnat}
\bibliography{literature_references}

\end{document}